\documentclass[fleqn,10pt]{wlscirep}
\usepackage[utf8]{inputenc}
\usepackage[T1]{fontenc}

\usepackage{amsmath,amsfonts}
\usepackage{algorithmic}
\usepackage{algorithm}
\usepackage{array}

\usepackage{textcomp}
\usepackage{stfloats}
\usepackage{url}
\usepackage{verbatim}
\usepackage{graphicx}
\usepackage{cite}

\usepackage[numbers]{natbib}
\usepackage{multicol}
\usepackage{hyperref}
\usepackage{bbding}

\usepackage{subfigure}
\usepackage{amssymb}
\usepackage{bm}
\usepackage{diagbox}
\usepackage{multirow}
\usepackage{ragged2e}
\usepackage{booktabs}
\usepackage{hhline}
\usepackage{colortbl}
\usepackage{caption}

\usepackage{pifont}

\usepackage{tabularray}
\usepackage[table]{xcolor}
\definecolor{Alto}{rgb}{0.878,0.878,0.878}

\usepackage{listings}
\title{Towards Task-Oriented Flying: Framework, Infrastructure, and Principles}

\author[1,*]{Kangyao Huang}
\author[2,*]{Hao Wang}
\author[3]{Jingyu Chen}
\author[1]{Jintao Chen}
\author[1]{Yu Luo}
\author[4]{Di Guo}
\author[2]{Xiangkui Zhang}
\author[1,$\dagger$]{Xiangyang Ji}
\author[1,$\dagger$]{Huaping Liu}
\affil[1]{Tsinghua University}
\affil[2]{Dalian University of Technology}
\affil[3]{Institue of Software, Chinese Academy of Sciences}
\affil[4]{Beijing University of Posts and Telecommunications}
\affil[*]{Contribute equally}
\affil[$\dagger$]{Corresponding author}

\begin{abstract}

Deploying robot learning methods to aerial robots in unstructured environments remains both challenging and promising. While recent advances in deep reinforcement learning (DRL) have enabled end-to-end flight control, the field still lacks systematic design guidelines and a unified infrastructure to support reproducible training and real-world deployment. We present a task-oriented framework for end-to-end DRL in quadrotors that integrates design principles for complex task specification and reveals the interdependencies among simulated task definition, training design principles, and physical deployment. Our framework involves software infrastructure, hardware platforms, and open-source firmware to support a full-stack learning infrastructure and workflow. Extensive empirical results demonstrate robust flight and sim-to-real generalization under real-world disturbances. By reducing the entry barrier for deploying learning-based controllers on aerial robots, our work lays a practical foundation for advancing autonomous flight in dynamic and unstructured environments.
\end{abstract}
\begin{document}

\flushbottom
\maketitle

\thispagestyle{empty}

\section{Introduction}

Aerial robots have emerged as revolutionary technologies that drive the development of the low-altitude economy\cite{Chu2024,Wuest2024}. Their applications span critical domains including emergency rescue operations, precision agriculture, and urban air mobility, with the boosting of global drone market. This explosive growth is fueled by advancements in autonomous capabilities that enable complex operations in unstructured environments. Recent advances in lightweight sensing technologies and edge computing further accelerate their deployment across industries.

Deep reinforcement learning (DRL) has demonstrated superior performance in time-critical tasks including drone racing\cite{Hanover2024,Kaufmann2023,Song2021, Nagami2021, Zhang2024a,Loquercio2020}, dynamic obstacle avoidance\cite{Singla2021,Xiao2023}, rapid trajectory tracking\cite{Chen2024,Ma2023,Ma2023a}, agile flight\cite{Wu2024, Zhang2024a, Heeg2024, Wang2024a}, as well as navigation and planning\cite{Xiao2023, Zhang2024a,Lee2024}, achieving unprecedented perception and decision-making efficiency. Although many studies have provided methods and skills\cite{Wang2024a,Heeg2024,Wu2024,Zhang2024a,Kaufmann2022,Azzam2024,Kalidas2023,Chen2024,Nagami2021,Ma2023} to improve algorithms and give experience on solving specific tasks, the principles of training and deploying a policy by DRL have not been summarized and refined, especially from a comprehensive and thorough perspective covering simulation, deployment, and their bridge. Most importantly, existing quadrotor works based on robot learning methods describe a lot about methodology but lack guidelines, principles, as well as underlying tools for implementation, and there are few open-source toolkits to facilitate the DRL training and transfer. These absences create significant barriers to practical application: researchers often struggle and feel confused about where to begin when designing and deploying a new task-oriented embodied flight task.

Our research aims to consolidate existing end-to-end DRL methods for aerial robots while incorporating our extensive experimentation and trials, distilling general task design techniques, and summarizing training and deployment guidelines for quadrotor DRL, thereby facilitating broader adoption by researchers.
Specifically, we refine principles for designing successful quadrotor DRL tasks and consolidate the factors that affect final performance during the sim-to-real process. Additionally, to support our work and facilitate broader research efforts, we propose a fully open-sourced and robust infrastructure and workflow for quadrotor DRL, including simulation environments, software, firmware, and hardware platforms. We also introduce comprehensive experiments across several different types of aerial robot learning tasks. Our proposed platform supports training and sim-to-real transfer in a remarkably short period of time, significantly accelerating algorithmic iteration.

\section{Results}

\begin{figure*}[t]
    \centering
    \includegraphics[width=\linewidth]{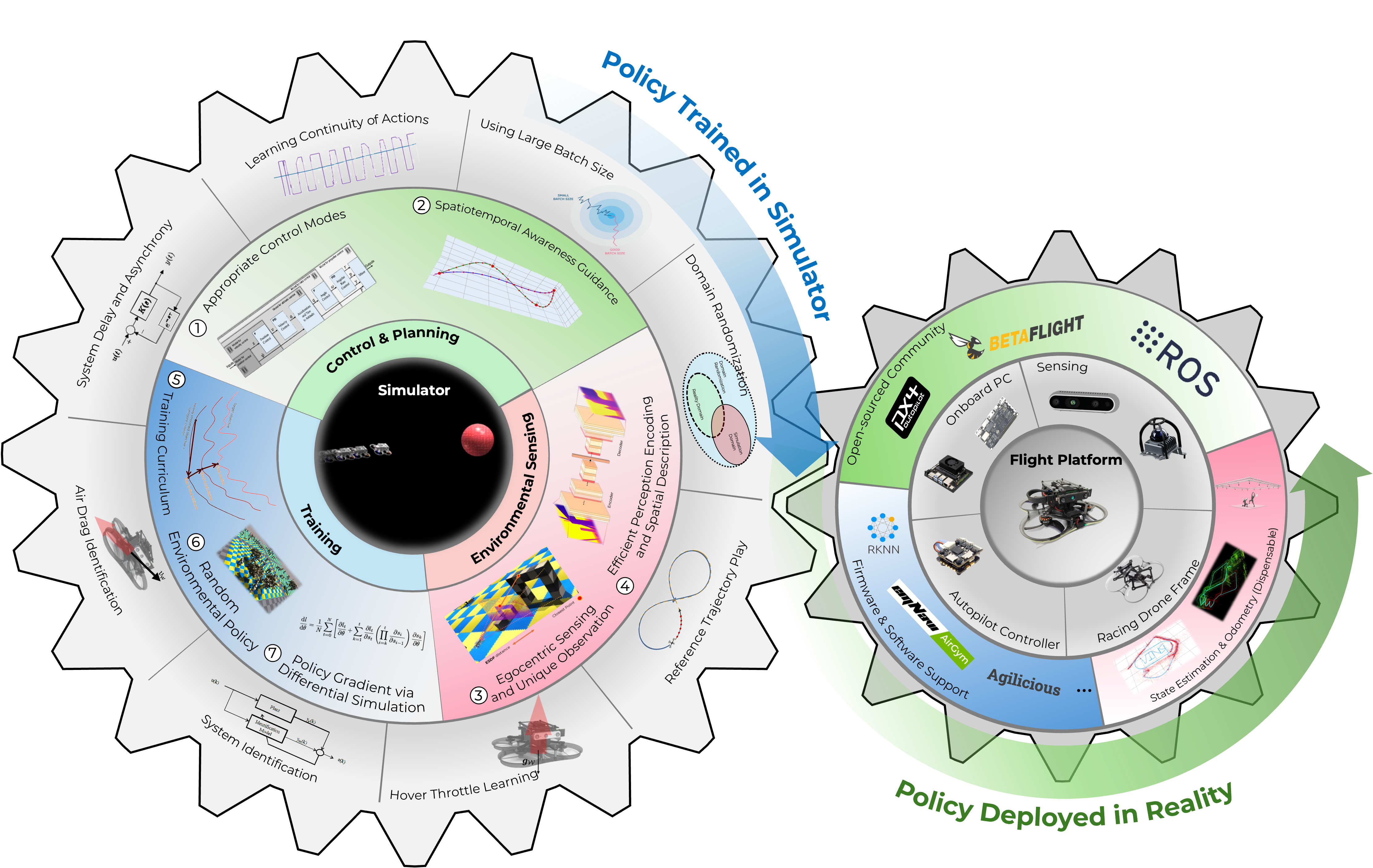}
    \caption{A DRL-based sim-to-real task-oriented framework. The left gear serves as the driving gear and represents simulation-related items, including simulator core, task principles middle layer, and sim-to-real techniques outermost layer. The right gear is the driven layer, involving hardware devices as inner layer, as well as firmware and software as outer layer.}
    \label{fig:gear}
\end{figure*}

\subsection{Framework}
The task-oriented framework is illustrated in Fig.\ref{fig:gear}. A pair of gears is employed as a metaphor to illustrate the interdependent relationship between simulation-based training and real-world deployment in aerial robot DRL. The left gear functions as the driving gear, symbolizing the training of policies within the simulator, while the smaller right gear serves as the driven gear, representing the deployment of these policies onto quadrotors in the real world. 

Simulator is the core of the driving gear, for training a policy. The interactive simulation environment, built upon physics engines and rendering systems, serves as the training arena for DRL. The physics engine simulates and approximates Newton's physical laws, forming the dynamical foundation for all complex flight behaviors. Concurrently, environmental perception and pattern recognition inherently depend on rendering verisimilitude, where graphical frame-by-frame progression drives the interactive learning process. Built upon the simulator, task-driven policy training requires systematic design considerations. While end-to-end quadrotor missions can encompass diverse task typologies, their design framework must address three fundamental dimensions: Control \& Planning, Environmental Sensing, and Training. This might involve multiple aspects, including command hierarchy selection, reward-punishment mechanisms, training environment design, and algorithm adaptation, each involving sophisticated techniques and diverse approaches. At this conceptual level, we systematically abstract and distill seven fundamental design principles addressing the aforementioned three dimensions, relying on both established literature and extensive empirical validation from our own practical works: \ding{172} appropriate control modes \ding{173} spatiotemporal awareness guidance \ding{174} egocentric sensing and unique observation \ding{175} efficient perception encoding and spatial description \ding{176} training curriculum \ding{177} random environmental policy \ding{178} policy gradient via differential simulation. The detailed explanation of these principles will be comprehensively elaborated later. The outermost layer of driving gear is factors and tricks that directly influence the performance of sim-to-real deployment. 

The driven gear can be conceptualized as the standard architecture of an autonomous unmanned system, embodying the following fundamental characteristics: platform, hardware device with firmware, and software application. As the validation platform for this framework and workflow, the self-developed aerial platform \texttt{X152b} is centrally positioned within the gear diagram. Hardware components such as the onboard PC, flight controller, and sensors are integrated into this platform. Sensing by vision\cite{Guerra2019,Loquercio2020,Amer2021,Ren2022,Wang2023a,Kulkarni2024,Xu2025,Zhang2024a,Foehn2022}, LiDAR\cite{Xu2025} and neuromorphic vision\cite{Delmerico2019,Hanover2024,Andersen2022} as well as other perception devices can also be employed. At the outermost layer, the associated firmware \& software, state estimation algorithms, and open-sourced communities serve as the medium for transferring and deploying a neural network onto the physical platform, and meshing with the outermost layer of driving gear. 

\subsection{Infrastructure \& Workflow}
To facilitate end-to-end policy training and deployment, we integrate insights from numerous existing RL studies along with our practical experience, and propose a general, task-oriented workflow for rapidly training and deploying DRL-based aerial robot flight policies.
The pipeline of our platform is illustrated in Fig. \ref{fig:pipeline}, where we have implemented two separate closed-loops for training and deployment, tailored for simulation and real-world environments respectively. Our workflow pipeline includes:
\begin{itemize}
    \item DRL simulation platform \texttt{AirGym}\footnote{\href{https://github.com/emNavi/AirGym}{\texttt{https://github.com/emNavi/AirGym}}}: a large-scale parallel environment simulation based on IsaacGym\cite{Makoviychuk2021}, where we implement four DRL tasks: \texttt{tracking}, \texttt{avoidance}, \texttt{target hitting}, \texttt{planning}, as well as the most basic flight control task \texttt{hovering}.
    \item Parallel flight controller \texttt{rlPx4Controller}\footnote{\href{https://github.com/emNavi/rlPx4Controller}{\texttt{https://github.com/emNavi/rlPx4Controller}}}: a parallel simulation flight geometric controller implemented in \texttt{C++} and encapsulated with \texttt{Python} interface, strictly aligned with the logic and parameters of the PX4 autopilot. It flexibly provides control at various levels, including position \& yaw ($PY$), linear velocity \& yaw ($LV$), collective thrust \& attitude angle ($CTA$), and collective thrust \& body rate ($CTBR$).
    \item Edge-side onboard inference runtime software \texttt{AirGym-Real}\footnote{\href{https://github.com/emNavi/AirGym-Real}{\texttt{https://github.com/emNavi/AirGym-Real}}}: an onboard sim-to-real module compatible with \texttt{AirGym}, enabling direct loading of pretrained models, supporting onboard visual-inertial pose estimation, \texttt{ROS} topic publishing, and one-shot scripted deployment.
    \item Control bridge \texttt{control\_for\_gym} \footnote{\href{https://github.com/emNavi/control_for_gym}{\texttt{https://github.com/emNavi/control\_for\_gym}}}: a middleware layer based on MAVROS for forwarding control commands at various levels to PX4 autopilot. It includes a finite state machine to facilitate switching between DRL models and traditional control algorithms.
    \item Quadrotor platform \texttt{X152b}: an open-sourced onboard sensing and inference flight platform, shown in Fig.\ref{fig:pipeline}, where we integrate computation and perception devices and provide the mechanical model for simulation alignment.
\end{itemize}

\begin{table*}[h!]
\centering
\caption{Time consumption for training a policy: training from scratch.}
\fontsize{7}{7}\selectfont
\renewcommand{\arraystretch}{1} 
\begin{tabular}{
    >{\centering\arraybackslash}p{2cm} 
    >{\centering\arraybackslash}p{2cm} 
    >{\centering\arraybackslash}p{2cm} 
    >{\centering\arraybackslash}p{2cm} 
    >{\centering\arraybackslash}p{2cm} 
    >{\centering\arraybackslash}p{2cm} 
}
\toprule
\textbf{Task Name} & \textbf{hovering} & \textbf{tracking} & \textbf{avoidance} & \textbf{target hitting} & \textbf{planning} \\
\midrule
Comsumption (min) & 2.4465 $\pm$ 0.1878 & 8.1619 $\pm$ 0.3811 & 5.3834 $\pm$ 0.4229 & 3.8277 $\pm$ 0.2125 & 7.1009 $\pm$ 0.3964 \\
\bottomrule
\end{tabular}
\label{tab:time}
\end{table*}

Our workflow demonstrates robust and rapid policy training-to-deployment capabilities, achieving mission-ready performance within minutes. As shown in Tab.\ref{tab:time}, the time consumptions of training a task in simulation among 10 trials are listed. In practice, the trained model is directly transferred to practical devices to realize zero-shot sim-to-real. Our training runs on a workstation with Intel Core i9-13900K processor and an NVIDIA RTX4090 graphic card, and the policy is deployed to an onboard computer with RK3588s processor. Since there is no collision during the training and flying, we can simply copy policies from the simulator to real-world robots without any fine-tuning which has been proved to be feasible in previous studies\cite{Song2021}. Additionally, our efficient workflow pipeline software offers one-click solutions for complex engineering implementations like robot state estimation and perception pre-processing, enabling immediate deployment and inference. Furthermore, the workflow is highly scalable; the modules and tools within the pipeline are not limited to the specific hardware platform we used. Features and usage of subsystems are described in supplementary materials.

\begin{figure*}[p]
    \centering
    \includegraphics[width=\textwidth, height=\textheight]{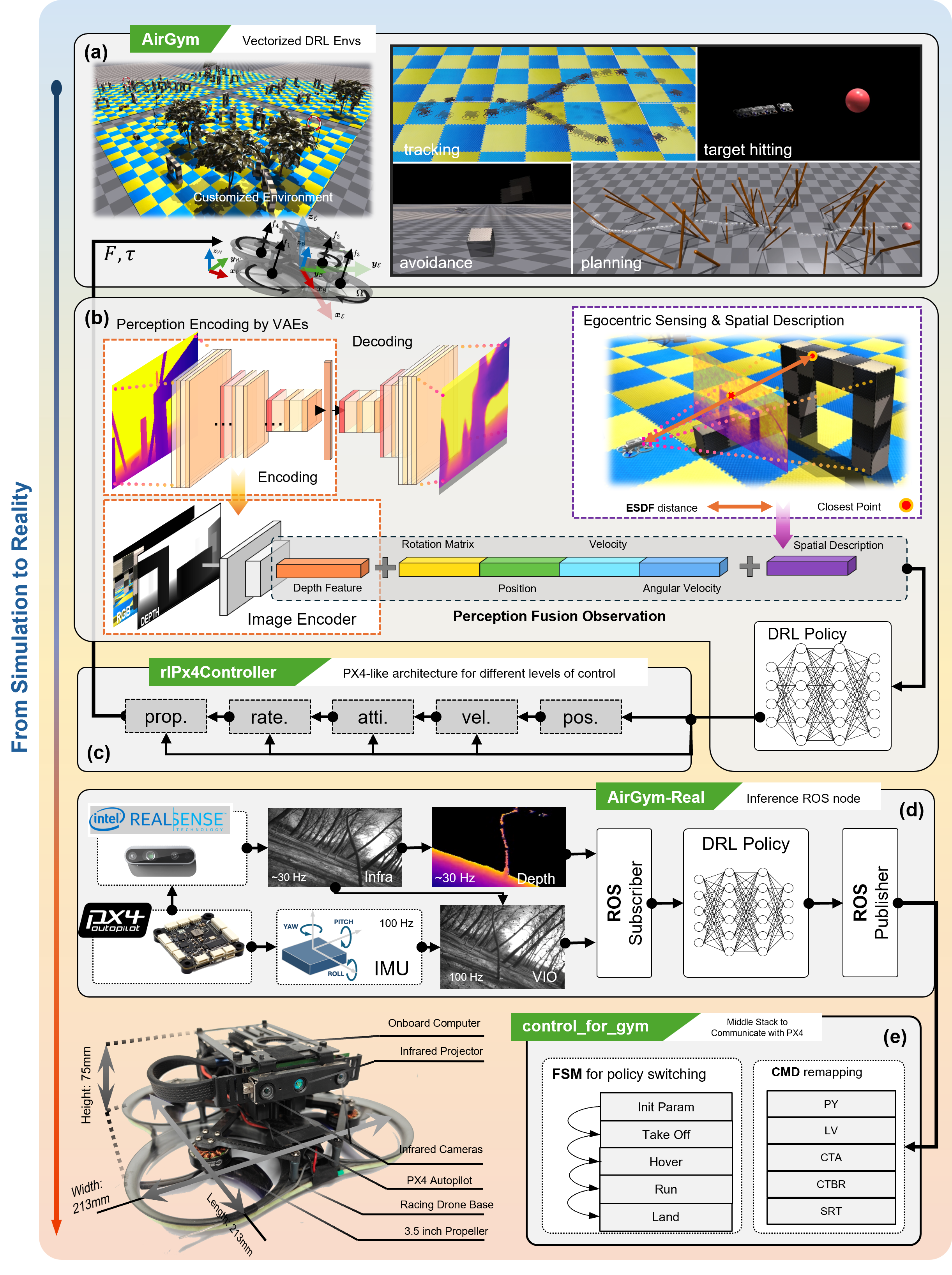}
\end{figure*}

\clearpage
\noindent
\begin{minipage}{\textwidth}
    \centering
    \captionof{figure}{The proposed pipeline and the relationships between each part. In (a) \texttt{AirGym}, we provide four classical tasks: \texttt{tracking}, \texttt{avoidance}, \texttt{target hitting}, and \texttt{planning}. (b) illustrates the sensing processing and features fusion during the training phase. The left image in (b) shows visual input is compressed into a vector using VAEs. The depth image is reconstructed by using self-supervised learning, and the encoding part is used as the image encoder in the DRL loop. The right image in (b) illustrates ESDF distance which is used as an efficient spatial description, where we use the minimum value in depth array is used as the ESDF distance. (c) shows all different control modes that can be selected. (d) is the onboard inference module that runs on the flight platform. It is worth mentioning that the VIO part is not indispensable in our experiments because global states estimation is not necessary in our Ego-centric design. (e) builds the bridge between PX4 flight controller and the algorithm. Finally, bottom-left device is our flight platform \texttt{X152b}.}
    \label{fig:pipeline}
\end{minipage}

\subsection{Principles in Training Design}
Advances in Sim-to-Real for aerial robots increasingly enable tackling more complex tasks. However, designing effective task-oriented DRL training remains challenging. Poorly defined training objectives may misguide the learner into blind alleys, reducing the problem to unsolvable reward engineering; conversely, well-crafted training acts as a scaffold, steering learning toward generalizable and robust policies. This paradigm shift necessitates a principled approach to task design. In the following, we outline the fundamental principles to guide effective training design.

\subsubsection{Control \& Planning}
The most fundamental question in learning-based control is deceptively simple: what exactly should the agent learn to control? The answer lies in selecting an appropriate level of abstraction for the control command—ranging from low-level motor thrusts to high-level position targets. Inspired by classical hierarchical control theory, which organizes control commands hierarchically, we adopt a similar strategy and define five distinct control modes: $PY$, $LV$, $CTA$, $CTBR$, and $SRT$. $CTBR$ emerges as a robust mode, striking a balance between expressiveness and tractability. It performs consistently well across a range of tasks, indicating strong generalization capabilities. Meanwhile, $LV$ mode demonstrates faster convergence in tasks with looser precision requirements, making it a practical alternative when high-frequency control is unnecessary. Notably, the choice of control mode directly influences shapes the agent’s action space and physical embodiment, shaping both the learning space and the physical embodiment of policy execution. Thus, it is crucial to \ding{172} \underline{select appropriate control mode} that matches the task at hand. For instance, while $SRT$ offers the most sensitive control, it poses significant learning challenges due to the complexity of the underlying dynamics. In contrast, mid-level abstractions such as $CTBR$ or $LV$ strike a more favorable balance between controllability and learning efficiency.

Planning complex trajectories poses significant challenges, while \ding{173} \underline{spatiotemporal awareness guidance} can facilitate it. Whether through incorporating desired poses as prior knowledge that embeds useful priors into the learning process\cite{Xu2024} or providing environmental spatial descriptors in the state space as a temporary target\cite{Song2021, Song2023}, both approaches can effectively enhance flexibility when handling complex trajectories. Whether these cues come from human commands or perception pipelines, their value lies in anchoring the agent’s decision-making in both space and time, allowing the drone to plan ahead—without relying on brittle rule-based controllers. Therefore, when an agent gets lost in the "reward engineering", it is important to rethink whether sufficient guidance has been provided to the agent, and whether this guidance has been properly incorporated into the state space.

\subsubsection{Environmental Sensing}
In most traditional and classical quadrotor control and navigation scenarios, researchers typically use (\romannumeral1) the takeoff point as the origin of coordinate and (\romannumeral2) adopt a North-East-Down (NED) coordinate frame. Because a locally constructed coordinate system like this allows the mathematical formulation of global/local path planning problems to be clearly defined from a third personal view, which is well-suited for control and planning approaches based on optimization but often hinders training in more embodied scenarios. The \ding{174} \underline{egocentric sensing}, where the drone perceives the world from its own point of view, enhancing learning stability and efficiency. We define an ego-centric coordinate system $\mathcal{E}$: the origin of this coordinate is fixed on the mass point of quadrotor, using the forward-facing direction of vehicle as the positive $\bm x_{\mathcal{E}}$. $\bm z_{\mathcal{E}}$ points vertically upward (toward the sky), and the $\bm y_{\mathcal{E}}$ is determined according to the right-hand rules. Ego-centric frame $(\bm x_{\mathcal{E}},\bm y_{\mathcal{E}},\bm z_{\mathcal{E}})$ is similar to the body frame $(\bm x_{\mathcal{B}},\bm y_{\mathcal{B}},\bm z_{\mathcal{B}})$ while the difference is $\bm z_{\mathcal{E}}$ in ego-centric frame always points to the sky. By anchoring perception to the drone’s forward-facing frame, the system avoids the confusion introduced by arbitrary global positions or orientations. In practice, designing the observation space with respect to the agent’s own perspective can greatly enhance learning stability and efficiency. This aligns not only with biological inspiration—humans and animals rely on self-centered perception—but also with practical benefits: observations become more compact, generalizable, and better suited for transfer to real-world deployments. Additionally, emphasize the importance of using unique state representations to ensure a one-to-one mapping from observations to actions. For example, while quaternions are commonly used for attitude representation. Since quaternions are not unique representations of the same rotation, we recommend using rotation matrices to represent orientation. This helps the neural network learn an exclusive mapping from input to output, and it makes the training converge faster.

Moreover, what a drone perceives is only as valuable as its ability to interpret it, which highlights the importance of \ding{175} \underline{efficient perception encoding and spatial description}. While raw depth images or point clouds offer rich environmental information, they are often too high-dimensional and noisy for direct use in learning policies. Instead, feature extraction—either through hand-crafted heuristics or learned encoders—can distill relevant information into compact latent variables. Probabilistic models like variational autoencoders (VAEs) are especially promising \cite{Kulkarni2024}. They can produce low-dimensional features that still preserve critical environmental cues. These encodings also offer smooth variations and a structured latent space, which are beneficial for stable training. When used with additional spatial priors, such as Euclidean Signed Distance Fields (ESDFs), they enable reward functions that naturally reflect obstacle proximity—encouraging safe and agile navigation \cite{Xu2025}. Therefore, both perception encoding and spatial description are indispensable: the former enhances the effective utilization of perceptual information, while the latter bridges perception and reward design.

\subsubsection{Training}
DRL training for drones must address the complexity of tasks. Many tasks involve intricate logical dependencies or require the integration of multiple skills that are hard to learn simultaneously, like dodging dynamic obstacles while hovering. A practical solution is to employ \ding{176} \underline{training curriculum}, such as deep hierarchical learning (DHL), to decompose the task into manageable sub-problems\cite{Hong2024}. For example, a system can first learn a perception module that maps depth inputs to compact feature representations, and then train a separate policy to act based on those features to improve training stability and overall performance. 

\ding{177} \underline{Random environmental policy} is essential for improving the generalization capability of learned policies. For example, such as tracking arbitrary trajectories, drone racing across various tracks, or navigating in unknown scenes, the ability of a model to adapt to diverse conditions is a fundamental requirement. Randomization strategies have been shown to be effective in the previous works, like environment policy in study \cite{Wang2024b} and the unified tracking policy in SimpleFlight \cite{Chen2024}.

In more complex tasks, \ding{178} \underline{differential simulation-based learning} yields more efficient and intelligent results\cite{Zhang2024a}. This approach involves designing the RL process—especially policy updates—as an end-to-end differentiable system, enabling the entire learning process to be efficiently optimized through gradient descent. Traditional RL often relies on sampling and the design of reward functions to obtain gradients, where the optimization objectives are often non-differentiable or suffer from high gradient estimation noise. However, differentiable reinforcement learning enables direct backpropagation of gradients, leading to lower estimation noise and more stable learning dynamics\cite{Heeg2024}. Therefore, rather than designing complex reward functions, it may be more effective to consider how to construct loss functions directly from differentiable components.

\subsection{Task-oriented Flying}
We implement four classic RL/DRL tasks and build a set of quadrotor experiments on top of \texttt{AirGym}: \texttt{tracking}, \texttt{avoidance}, \texttt{target hitting}, and \texttt{planning}, as shown in Fig.\ref{fig:pipeline}(a). All tasks support end-to-end policy training across various control levels, allowing us to select and integrate specific levels where we wish to train and maintain classical control methods.  \texttt{tracking} is a classical drone robot learning task: the quadrotor follows the given trajectory with a desired speed. Herein we primarily conduct tracking of a lemniscate trajectory flight. Task \texttt{avoidance} is a new scenario not addressed in previous RL tasks: the quadrotor starts at a fixed point, then an obstacle will be initialized at a random position with an appropriate speed to hit the quadrotor. The quadrotor needs to learn a policy to avoid the obstacle and keep hovering at the original position. \texttt{target hitting} is a waypoint navigation task: the quadrotor flies at a very high speed, rapidly dashing toward the target. Finally, we propose a \texttt{planning} task: a cluster of trunks is randomly generated at the beginning of an episode, and a quadrotor needs to fly through the forest and reach the target.

In our experiments, we evaluate the results of all different control types. The $CTBR$ control mode is recommended as its effectiveness has been studied and shown to outperform others \cite{Kaufmann2022, Xu2024}. Our experimental results also support this perspective: although $CTBR$ may not offer the fastest training speed or the highest reward, its flight behavior is more stable and natural, based on real-world observations. Therefore, we uniformly use the $CTBR$ mode for sim-to-real experiments.

\subsubsection{Tracking with Unknown Disturbance}


\begin{figure*}[p]
    \includegraphics[width=\textwidth]{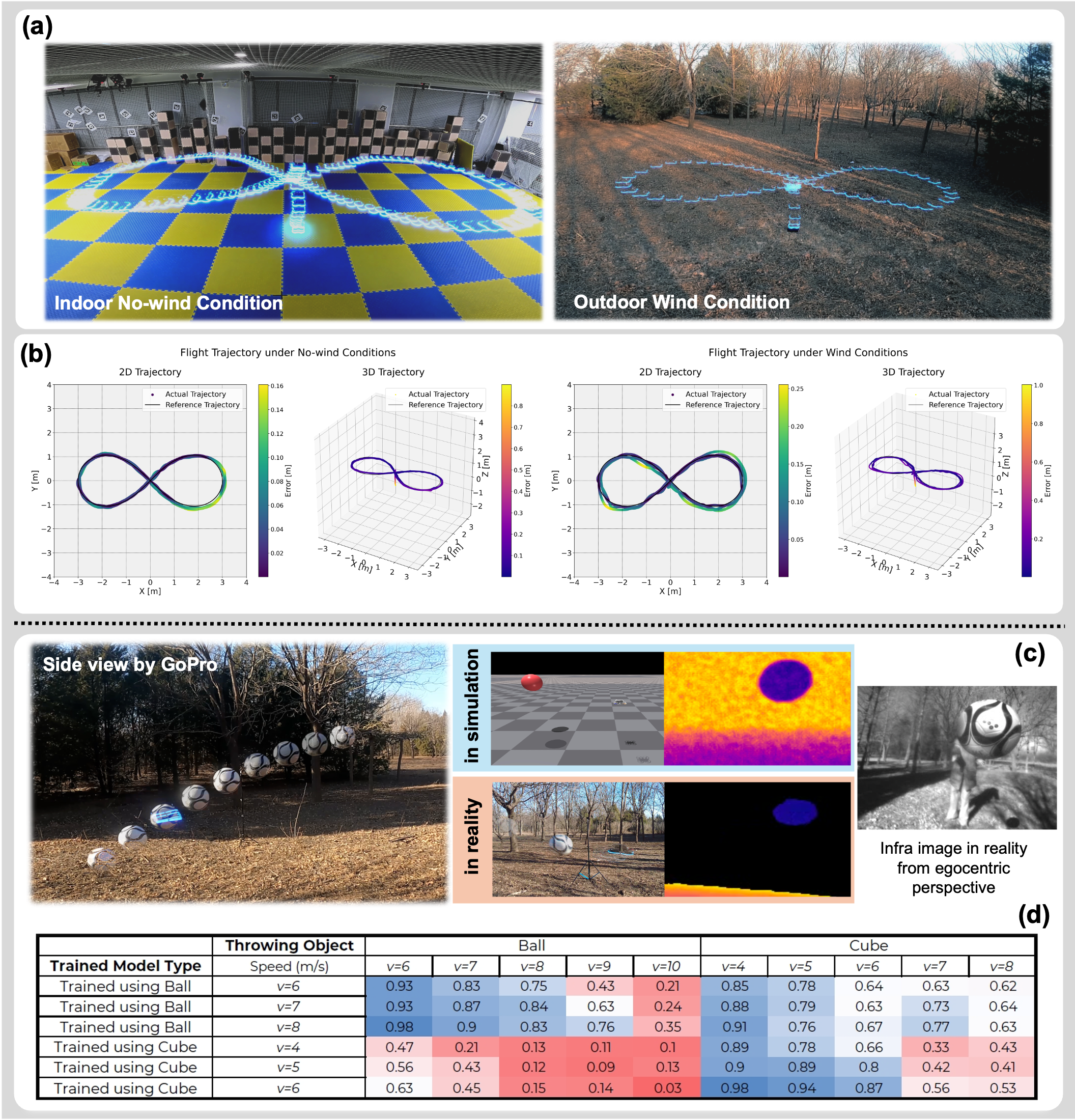}
    \caption{(a)(b) show the tracking performance under different conditions. We set a lemniscate trajectory as tracking reference and illustrate the cumulative flight path under indoor no-wind and outdoor windy environments. (c) is dynamic obstacle avoidance task by end-to-end DRL Sim-to-Real using depth sensing. The quadrotor with blue halo tries to dodge a fast flying football. Our experiment is conducted in a wild GNSS-deny environment and only relies on onboard sensing for perception and localization. The speed of flying ball reaches about $15 \mathrm{m\cdot s^{-1}}$. (d) shows the generalization experiments by using different throwing objects with various speed. One hundred episodes are evaluated for each experiment and success rate are recorded.}
    \label{fig:track-avoid}
\end{figure*}

Tracking a lemniscate is a commonly used baseline for demonstrating control performance. Here, we conduct lemniscate tracking experiments under both ideal indoor and dim, windy outdoor conditions, as shown in Fig.\ref{fig:track-avoid}(a)(b) respectively. Meanwhile, we present the tracking performance at two different tracking speeds.

\begin{table}
\centering
\caption{Comparison between different conditions of lemniscate tracking.}
\label{tab:tracking}
\fontsize{6}{6}\selectfont
\begin{tblr}{
  width = \linewidth,
  colspec = {Q[327]Q[94]Q[327]Q[190]},
  cells = {c},
  cell{2}{1} = {r=2}{},
  cell{2}{4} = {Alto},
  cell{3}{4} = {Alto},
  cell{4}{1} = {r=2}{},
  cell{4}{4} = {Alto},
  cell{5}{4} = {Alto},
  cell{6}{1} = {r=2}{},
  cell{6}{4} = {Alto},
  cell{7}{4} = {Alto},
  cell{8}{1} = {r=2}{},
  cell{8}{4} = {Alto},
  cell{9}{4} = {Alto},
  cell{10}{1} = {r=2}{},
  cell{10}{4} = {Alto},
  cell{11}{4} = {Alto},
  vlines = {0.05em},
  hline{1,12} = {-}{0.08em},
  hline{2} = {1-2}{0.08em},
  hline{2} = {3-4}{0.03em},
  hline{3,5,7,9,11} = {2-4}{0.03em},
  hline{4,6,8,10} = {-}{0.05em},
}
\textbf{Items}                   & \textbf{Speed} & \textbf{Relative Tracking Errors} & \textbf{Success Rate} \\
SimpleFlight (Baseline)          & 1.6m/s         & 0.028 $\pm$ 0.0                        & –                     \\
                                 & 2.5m/s         & 0.051 $\pm$ 0.002                      & –                     \\
\textbf{Indoor with no wind}     & 1.6m/s         & 0.022 $\pm$ 0.013  $\searrow$           & 1                     \\
                                 & 2.5m/s         & 0.03 $\pm$ 0.006  $\searrow$            & 0.9                   \\
\textbf{Outdoor with no wind}    & 1.6m/s         & 0.023 $\pm$ 0.015 $\searrow$           & 1                     \\
                                 & 2.5m/s         & 0.032 $\pm$ 0.024 $\searrow$            & 0.8                   \\
\textbf{Outdoor with wind}       & 1.6m/s         & 0.073 $\pm$ 0.043                      & 0.3                   \\
                                 & 2.5m/s         & 0.11 $\pm$ 0.041                       & 0.2                   \\
\textbf{Outdoor with wind \& DR} & 1.6m/s         & 0.043 $\pm$ 0.028 $\searrow$                     & 0.6 $\nearrow$                    \\
                                 & 2.5m/s         & 0.082 $\pm$ 0.035 $\searrow$                     & 0.5 $\nearrow$                    
\end{tblr}
\end{table}

As illustrated in Fig. \ref{fig:track-avoid}(b), we use the Mean Euclidean Distance (MED) to evaluate the tracking errors. The results report a maximum $0.09\pm 0.07$ meters of MED under no-wind conditions. However, tracking lemniscate under moderate breeze conditions presents significant challenges. The tracking performance is shown in Fig. \ref{fig:track-avoid}(b). Because the wind direction and strength in the outdoor environment are both uncertain, the disturbances can be considered approximately random. Thus, the trajectory is irregular, and it deviates from the reference for most of the process. The wind disturbance directly affects the quadrotor's attitude, decreasing the control accuracy of the neural network. On the other hand, persistent wind disturbances may disrupt the quadrotor's control, making it difficult to follow up the desired playing trajectory and leading to a crash. When delay or ahead of trajectory playback occurs, the neural network encounters states that it has never seen during training, which can lead to unpredictable outcomes \cite{Gleave2020a}. To conquer this, domain randomization can significantly improve the robustness, as shown in Tab.\ref{tab:tracking}, by adding the temporal margin \cite{Chen2024} and randomization on observation. Furthermore, artificially adding Gaussian-distributed external force disturbances to simulate wind as a kind of DR method can also increase the success rate. The results show that the success rate nearly doubles after DR is applied.

To make the results more persuasive, we make a comparison with SimpleFlight\cite{Chen2024}. Due to the different scales of flight platforms, we use the relative proportion of tracking error for a fairer comparison. Note that the maximum edge length of the trajectory in SimpleFlight is limited to $(-1, 1)$, whereas in our experiment it reaches $(-3, 3)$. In our experiment, the relative tracking error achieves a lower average compared to the baseline, as shown in Tab.~\ref{tab:tracking}, which demonstrates the effectiveness of our method in outdoor environments. Moreover, we apply DR to the timing of sending the desired tracking trajectory by introducing slight delays and random time offsets $T\sim\mathcal{N}(\mu,\sigma^2 )$, where $(\mu,\sigma)=(0.3,0.5)$. Compared with the experiments without DR under windy conditions, the incorporation of DR effectively improves the success rate of the tracking experiments and reduces the relative tracking errors. It is worth mentioning that the variance of our method is higher than the baseline because the observation in our experiments comes from VIO while an external optical positioning system in SimpleFlight. Although VIO drift and quadrotor vibrations may increase the variance in tracking performance, they do not affect the overall tracking trend, as our tracking error is also computed relative to the position data from VIO.

\subsubsection{Dynamic Obstacle Avoidance}
The experiments above already demonstrate the system's Sim-to-Real stability for routine tasks. In this experiment, we aim to highlight the end-to-end inference capability when faced with rapidly changing environments. As illustrated in Fig. \ref{fig:track-avoid}(c), the quadrotor has to avoid a rapidly flying object, a ball or a cube assembled from EVA foam mats, thrown from $6$ meters away. The speed of the flying ball is $10\sim15$ m/s. Fig.\ref{fig:track-avoid}(c) is the processed depth image from the camera, where depths beyond $4.5$ meters are truncated. Considering the high speed of the obstacle, it places high demands on the real-time neural network inference. The success rate of dynamic obstacle avoidance reaches over $90\%$ across more than 20 trials involving either a ball or a cube.

To evaluate the generalization capability of our end-to-end model, we conduct transfer experiments by training models with two different types of dynamic obstacles—ball and cube—each at distinct throwing speeds, and subsequently testing them under varying conditions. As illustrated in Fig.\ref{fig:track-avoid}(d), the training involved dynamic obstacles moving at speeds ranging from $4\;\mathrm{m\cdot s^{-1}}$ to $8\;\mathrm{m\cdot s^{-1}}$. We perform cross-testing using models trained under different configurations to assess performance when encountering previously unseen obstacle dynamics. Additionally, we increase the maximum velocity of incoming obstacles during testing to examine the robustness of the learned policies in more extreme scenarios. The results indicate that the trained model exhibits a certain level of robustness when dealing with dynamic objects of unknown types. The model trained with balls as dynamic obstacles demonstrates better generalization capability, likely because the balls move faster than cubes during training. As a result, the trained model can more easily adapt its motion planning strategy to handle high-speed objects.

\subsubsection{High-speed Flight}
\begin{figure*}[t]
    \centering
    \includegraphics[width=\linewidth]{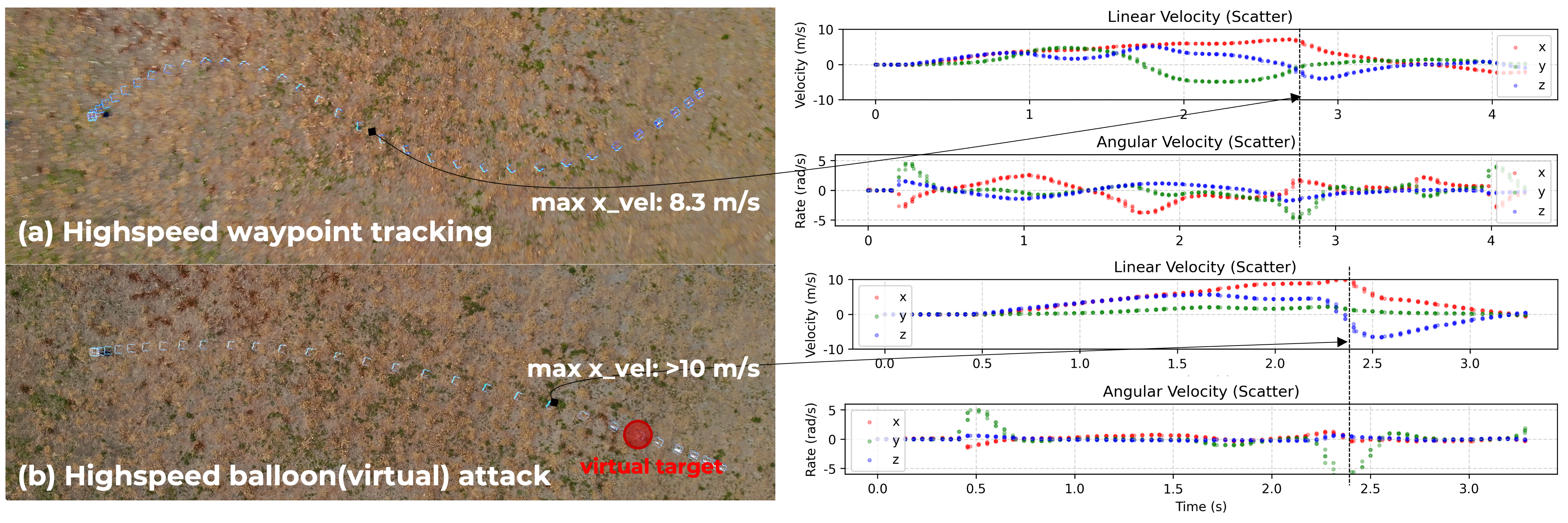}
    \caption{High-speed target hitting results. The blue halo is the quadrotor \texttt{X152b} with a strip light. (a) shows the high speed tracking with a sigmoid reference. (b) shows a \texttt{target hitting} task that accelerates from $0\mathrm{m\cdot s^{-1}}$ and hits a virtual target.}
    \label{fig:high-speed}
\end{figure*}
In order to evaluate the performance of our workflow in dealing with extreme maneuvers, we conduct a series of high-speed impact and tracking experiments based on the \texttt{target hitting} and \texttt{tracking} tasks, as illustrated in Fig.\ref{fig:high-speed}. The blue trajectory demonstrates the high-speed flight path. Here we test the sim-to-real performance on both the \texttt{target hitting} and \texttt{tracking} tasks. In the \texttt{target hitting} task, a virtual balloon is placed at a fixed position, and the quadrotor must accelerate from rest to strike the target. Furthermore, in the \texttt{tracking} task, the quadrotor follows a sigmoid-shaped trajectory starting from rest. In both experiments, no upper speed limit is imposed, allowing the quadrotor to reach the maximum velocity it can learn in \texttt{AirGym}. Linear and angular velocity records are shown on the right side of Fig.\ref{fig:high-speed}. Remarkably, the platform achieves a top speed exceeding $10\;\mathrm{m\cdot s^{-1}}$, demonstrating its strong maneuverability.

\subsubsection{Navigation in Forest}
\begin{figure*}
    \centering
    \includegraphics[width=\linewidth]{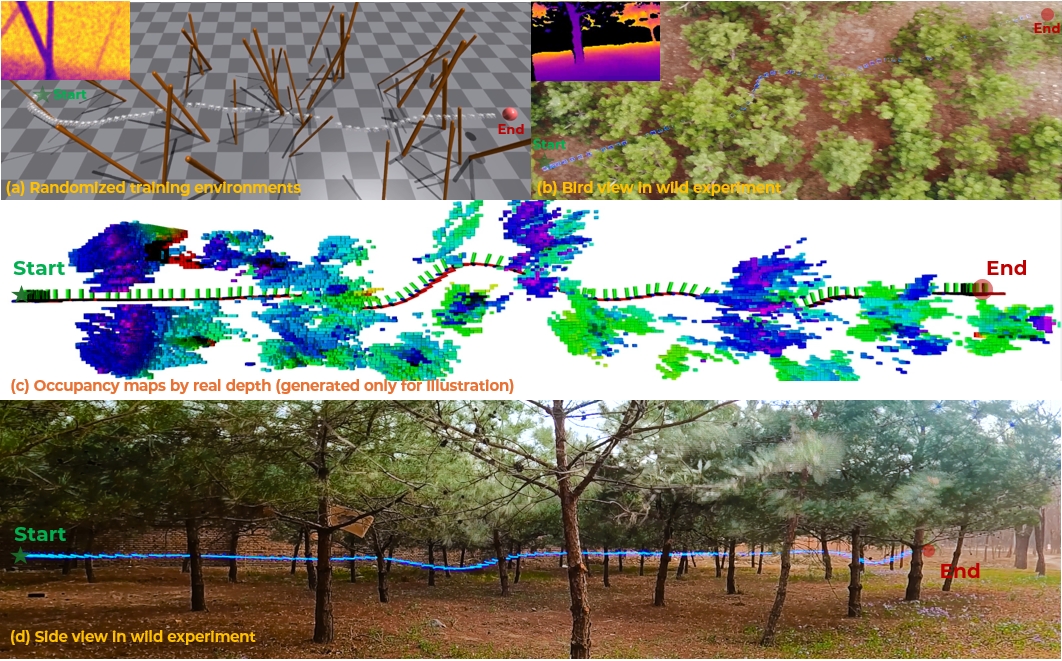}
    \caption{Navigating to fly out of the woods by task \texttt{Planning} sim-to-real. The sub-figure at the right bottom is generated using depth image after experiments and only for terrain illustration.}
    \label{fig:planning}
\end{figure*}
We set up an outdoor wild flight scenario where the quadrotor navigates through and out of a small forest, illustrated in Fig.\ref{fig:planning}. Herein, we evaluate the effectiveness of policies trained in the \texttt{planning} task. The quadrotor is initialized at the start point and then flies to the end with $1.5\;\mathrm{m\cdot s^{-1}}$ maximum speed. The quadrotor navigates using depth images as input. Since the branches in the wild are dense and chaotic, differing somewhat from the simulation scene, we strictly constrain the actual flight altitude to around 0.5 meters to adjust the field of view (FOV) to prevent the camera from capturing images of canopies and other elements not present in the simulation. Moreover, we clip the depth image range to $0\sim4.5$ meters to further reduce the likelihood of capturing canopies. The quadrotor autonomously navigates a distance of over $20$ meters and successfully reaches the target location as shown in the bird’s-eye view Fig.\ref{fig:planning}(b). In Fig.\ref{fig:planning}(c), we use the recorded depth stream to generate the occupancy map of the area, highlighting the complexity of the forest environment.

\section{Discussion}
The direction of robot learning has emerged as a transformative force in aerial robotics, enabling autonomous systems to adapt, learn, and perform complex tasks in dynamic environments. This transformation goes beyond the conventional approach of merely grafting traditional pattern recognition methods onto aerial robot platforms to enhance onboard capabilities. Instead, it fundamentally redefines the control, planning, and perception architecture of quadrotor from the ground up, aiming to develop one task-driven end-to-end policy to realize all functions. This trend has sparked the emergence of increasingly challenging tasks in aerial robotics, as evidenced by the surge of related research in recent years\cite{Chen2024,Llanes2024,Azzam2024,Loquercio2020,Kaufmann2022,Panerati2021,Kaufmann2023,Hwangbo2017,Kulkarni2024,Ma2023,Foehn2022,Loquercio2020,Loquercio2021,Ma2023a,Heeg2024,Wang2024a,Song2021,Kalidas2023,Xiao2023,Lin2018,Zhou2021,Chen2023a,Song2023,Wu2024,Kulkarni2023,Song2020,Singla2021,Miera2023,Xu2024,Guerra2019,Wang2024b,Moon2019,Mohta2018,Foehn2020,Guerra2019a,Sun2021a,Romero2022,Penicka2022,Zhou2021b,Muller2019,Fu2023,Zhang2023e,Amer2021,Kulkarni2024,Yu2024,Lamberti2024,Pham2022,Tordesillas2023,Panerati2021b,Bajcsy2024}. In an increasing number of complex tasks, traditional methods are being replaced by end-to-end learning-based approaches, which serves as strong evidence of the generalization and effectiveness of learning-based methods. Among these, the majority of approaches are based on RL and its related variants. 

\begin{table*}[h!]
\centering
\caption{Comparison of previous highlight works in learning-based aerial robotics field.}
\fontsize{7}{7}\selectfont
\renewcommand{\arraystretch}{1} 
\begin{tabular}{
    >{\centering\arraybackslash}p{3.6cm} 
    >{\centering\arraybackslash}p{0.5cm} 
    >{\centering\arraybackslash}p{1cm} 
    >{\centering\arraybackslash}p{1cm} 
    >{\centering\arraybackslash}p{2cm} 
    >{\centering\arraybackslash}p{0.2cm} 
    >{\centering\arraybackslash}p{0.3cm} 
    >{\centering\arraybackslash}p{0.2cm} 
    >{\centering\arraybackslash}p{0.2cm} 
    >{\centering\arraybackslash}p{0.2cm}
    >{\centering\arraybackslash}p{2cm}
    >{\centering\arraybackslash}p{1cm}
}
\toprule
\textbf{Work} & \textbf{Release} & \textbf{Type} & \textbf{Method} & \textbf{Tasks} & \textbf{SRT} & \textbf{CTBR} & \textbf{CTA} & \textbf{LV} & \textbf{PY}/Traj & \textbf{Controller} & \textbf{Principles} \\
\midrule
Control of a quadrotor with reinforcement learning\cite{Hwangbo2017} & 2017 & Methodology & RL & Control & \textcolor[RGB]{50,205,50}{\Checkmark} & \textcolor[RGB]{139,0,0}{\XSolidBrush} & \textcolor[RGB]{139,0,0}{\XSolidBrush} & \textcolor[RGB]{139,0,0}{\XSolidBrush} & \textcolor[RGB]{139,0,0}{\XSolidBrush} & — & \ding{174} \\
\midrule
FlightGoggles\cite{Guerra2019} & 2019 & Simulator & — & Drone Racing & — & — & — & — & — & — & — \\
\midrule
Deep drone racing\cite{Loquercio2020} & 2020 & Methodology & DL & Drone Racing & \textcolor[RGB]{139,0,0}{\XSolidBrush} & \textcolor[RGB]{139,0,0}{\XSolidBrush} & \textcolor[RGB]{139,0,0}{\XSolidBrush} & \textcolor[RGB]{139,0,0}{\XSolidBrush} & \textcolor[RGB]{50,205,50}{\Checkmark} & —  & \ding{175}\\
\midrule
Flightmare\cite{Song2020} & 2020 & Simulator & RL & Control / Drone Racing & \textcolor[RGB]{50,205,50}{\Checkmark} & \textcolor[RGB]{50,205,50}{\Checkmark} & \textcolor[RGB]{139,0,0}{\XSolidBrush} & \textcolor[RGB]{139,0,0}{\XSolidBrush} & \textcolor[RGB]{139,0,0}{\XSolidBrush} & —  & \ding{172}\\
\midrule
NeuroBEM\cite{Bauersfeld2021} & 2021 & Dataset & DL & Agile Flight / Control & — & — & — & — & — & Betaflight  & — \\
\midrule
gym-pybullet-drones\cite{Panerati2021} & 2021 & Simulator & RL & Control / Multi-agent Lead-follower & \textcolor[RGB]{50,205,50}{\Checkmark} & \textcolor[RGB]{139,0,0}{\XSolidBrush} & \textcolor[RGB]{139,0,0}{\XSolidBrush} & \textcolor[RGB]{50,205,50}{\Checkmark} & \textcolor[RGB]{139,0,0}{\XSolidBrush} & —  & \ding{172}\\
\midrule
Learning high-speed flight in the wild\cite{Loquercio2021} & 2021 & Methodology & IL & Planning & \textcolor[RGB]{139,0,0}{\XSolidBrush} & \textcolor[RGB]{139,0,0}{\XSolidBrush} & \textcolor[RGB]{139,0,0}{\XSolidBrush} & \textcolor[RGB]{139,0,0}{\XSolidBrush} & \textcolor[RGB]{50,205,50}{\Checkmark} & MPC  & \ding{173}\ding{174}\ding{177}\\
\midrule
Autonomous drone racing with deep reinforcement learning\cite{Song2021} & 2021 & Methodology & RL & Drone Racing & \textcolor[RGB]{50,205,50}{\Checkmark} & \textcolor[RGB]{139,0,0}{\XSolidBrush} & \textcolor[RGB]{139,0,0}{\XSolidBrush} & \textcolor[RGB]{139,0,0}{\XSolidBrush} & \textcolor[RGB]{139,0,0}{\XSolidBrush} & Betaflight  & \ding{174}\ding{175}\ding{177}\\
\midrule
Agilicious\cite{Foehn2022} & 2022 & Software & — & Infrastructure for Any Tasks & \textcolor[RGB]{50,205,50}{\Checkmark} & \textcolor[RGB]{50,205,50}{\Checkmark} & \textcolor[RGB]{50,205,50}{\Checkmark} & \textcolor[RGB]{50,205,50}{\Checkmark} & \textcolor[RGB]{50,205,50}{\Checkmark} & Betaflight  & \ding{172}\\
\midrule
Aggressive quadrotor flight using curiosity-driven reinforcement learning\cite{Sun2022b} & 2022 & Methodology & RL & Planning & \textcolor[RGB]{139,0,0}{\XSolidBrush} & \textcolor[RGB]{139,0,0}{\XSolidBrush} & \textcolor[RGB]{50,205,50}{\Checkmark} & \textcolor[RGB]{139,0,0}{\XSolidBrush} & \textcolor[RGB]{139,0,0}{\XSolidBrush} & AscTech Autopilot & \ding{175}\ding{176}\\
\midrule
A deep multi-agent reinforcement learning framework for autonomous aerial navigation to grasping points on loads\cite{Chen2023a} & 2023 & Methodology & RL & Multi-agent Transport & \textcolor[RGB]{139,0,0}{\XSolidBrush} & \textcolor[RGB]{139,0,0}{\XSolidBrush} & \textcolor[RGB]{139,0,0}{\XSolidBrush} & \textcolor[RGB]{50,205,50}{\Checkmark} & \textcolor[RGB]{139,0,0}{\XSolidBrush} & Crazyflie Controller  & —\\
\midrule
OmniDrones\cite{Xu2024} & 2023 & Simulator & RL & Control / Tracking / Transport / Formation & \textcolor[RGB]{50,205,50}{\Checkmark} & \textcolor[RGB]{50,205,50}{\Checkmark} & \textcolor[RGB]{50,205,50}{\Checkmark} & \textcolor[RGB]{50,205,50}{\Checkmark} & \textcolor[RGB]{50,205,50}{\Checkmark} & —  & \ding{172}\\
\midrule
Champion-level drone racing using deep reinforcement learning\cite{Kaufmann2023} & 2023 & Methodology & RL & Drone Racing & \textcolor[RGB]{139,0,0}{\XSolidBrush} & \textcolor[RGB]{50,205,50}{\Checkmark} & \textcolor[RGB]{139,0,0}{\XSolidBrush} & \textcolor[RGB]{139,0,0}{\XSolidBrush} & \textcolor[RGB]{139,0,0}{\XSolidBrush} & Betaflight  & \ding{174}\ding{175}\\
\midrule
Reaching the limit in autonomous racing\cite{Song2023} & 2023 & Methodology & RL & Drone Racing & \textcolor[RGB]{139,0,0}{\XSolidBrush} & \textcolor[RGB]{50,205,50}{\Checkmark} & \textcolor[RGB]{139,0,0}{\XSolidBrush} & \textcolor[RGB]{139,0,0}{\XSolidBrush} & \textcolor[RGB]{139,0,0}{\XSolidBrush} & Betaflight  & \ding{174}\ding{175}\ding{177}\\
\midrule
Aerial gym simulator\cite{Kulkarni2023} & 2023 & Simulator & RL & Control / Planning & \textcolor[RGB]{139,0,0}{\XSolidBrush} & \textcolor[RGB]{139,0,0}{\XSolidBrush} & \textcolor[RGB]{50,205,50}{\Checkmark} & \textcolor[RGB]{50,205,50}{\Checkmark} & \textcolor[RGB]{50,205,50}{\Checkmark} & —  & \ding{174}\\
\midrule
Deep-PANTHER\cite{Tordesillas2023} & 2023 & Methodology & RL & Planning & \textcolor[RGB]{139,0,0}{\XSolidBrush} & \textcolor[RGB]{139,0,0}{\XSolidBrush} & \textcolor[RGB]{139,0,0}{\XSolidBrush} & \textcolor[RGB]{139,0,0}{\XSolidBrush} & \textcolor[RGB]{50,205,50}{\Checkmark} & —  & \ding{173}\\
\midrule
Learning quadrotor control from visual features using differentiable simulation\cite{Heeg2024} & 2024 & Methodology & RL & Control & \textcolor[RGB]{139,0,0}{\XSolidBrush} & \textcolor[RGB]{50,205,50}{\Checkmark} & \textcolor[RGB]{139,0,0}{\XSolidBrush} & \textcolor[RGB]{139,0,0}{\XSolidBrush} & \textcolor[RGB]{139,0,0}{\XSolidBrush} & Betaflight  & \ding{178}\\
\midrule
Back to Newton's law\cite{Zhang2024a} & 2024 & Methodology & RL & (Multi-agent) Planning & \textcolor[RGB]{50,205,50}{\Checkmark} & \textcolor[RGB]{139,0,0}{\XSolidBrush} & \textcolor[RGB]{139,0,0}{\XSolidBrush} & \textcolor[RGB]{139,0,0}{\XSolidBrush} & \textcolor[RGB]{139,0,0}{\XSolidBrush} & Betaflight/PX4  & \ding{174}\ding{177}\ding{178}\\
\midrule
What matters RL sim2real\cite{Chen2024} & 2024 & Methodology & RL & Control / Tracking & \textcolor[RGB]{139,0,0}{\XSolidBrush} & \textcolor[RGB]{50,205,50}{\Checkmark} & \textcolor[RGB]{139,0,0}{\XSolidBrush} & \textcolor[RGB]{139,0,0}{\XSolidBrush} & \textcolor[RGB]{139,0,0}{\XSolidBrush} & Crazyflie Controller  & \ding{172}\ding{173}\ding{177}\\
\midrule
Dashing for the golden snitch\cite{Wang2024a} & 2024 & Methodology & RL & Multi-agent Tracking & \textcolor[RGB]{139,0,0}{\XSolidBrush} & \textcolor[RGB]{50,205,50}{\Checkmark} & \textcolor[RGB]{139,0,0}{\XSolidBrush} & \textcolor[RGB]{139,0,0}{\XSolidBrush} & \textcolor[RGB]{139,0,0}{\XSolidBrush} & Betaflight  & —\\
\midrule
Reinforcement learning for collision-free flight exploiting deep collision encoding\cite{Kulkarni2024} & 2024 & Methodology & RL & Planning & \textcolor[RGB]{139,0,0}{\XSolidBrush} & \textcolor[RGB]{139,0,0}{\XSolidBrush} & \textcolor[RGB]{139,0,0}{\XSolidBrush} & \textcolor[RGB]{50,205,50}{\Checkmark} & \textcolor[RGB]{139,0,0}{\XSolidBrush} & PixRacer  & \ding{174}\ding{175}\ding{176}\\
\midrule
Learning-based navigation and collision avoidance through reinforcement for UAVs\cite{Azzam2024} & 2024 & Methodology & RL & Planning / Avoidance & \textcolor[RGB]{139,0,0}{\XSolidBrush} & \textcolor[RGB]{139,0,0}{\XSolidBrush} & \textcolor[RGB]{139,0,0}{\XSolidBrush} & \textcolor[RGB]{50,205,50}{\Checkmark} & \textcolor[RGB]{139,0,0}{\XSolidBrush} & —  & \ding{176} \\
\midrule
Flying on point clouds with reinforcement learning\cite{Xu2025} & 2025 & Methodology & RL & Planning & \textcolor[RGB]{139,0,0}{\XSolidBrush} & \textcolor[RGB]{50,205,50}{\Checkmark} & \textcolor[RGB]{139,0,0}{\XSolidBrush} & \textcolor[RGB]{139,0,0}{\XSolidBrush} & \textcolor[RGB]{139,0,0}{\XSolidBrush} & PX4  & \ding{174}\ding{175}\ding{177}\\
\midrule
NavRL\cite{Xu2024a} & 2025 & Methodology & RL & Planning & \textcolor[RGB]{139,0,0}{\XSolidBrush} & \textcolor[RGB]{139,0,0}{\XSolidBrush} & \textcolor[RGB]{139,0,0}{\XSolidBrush} & \textcolor[RGB]{50,205,50}{\Checkmark} & \textcolor[RGB]{139,0,0}{\XSolidBrush} & PX4  & \ding{174}\ding{175}\ding{177}\\
\midrule
\textbf{AirGym} (Ours) & 2025 & Workflow & RL & Control / Tracking / Target Hitting / Avoidance / Planning / Customized & \textcolor[RGB]{50,205,50}{\Checkmark} & \textcolor[RGB]{50,205,50}{\Checkmark} & \textcolor[RGB]{50,205,50}{\Checkmark} & \textcolor[RGB]{50,205,50}{\Checkmark} & \textcolor[RGB]{50,205,50}{\Checkmark} & PX4/rlPx4controller  & \ding{172}\ding{173}\ding{174}\ding{175}\ding{176}\ding{177}\\
\bottomrule
\end{tabular}
\label{tab:comparison}
\end{table*}

Despite the numerous learning-based methods that have been successfully implemented, a significant challenge for many newcomers to the field lies in the complexity of building and implementing such system-level engineering solutions. Moreover, the reproduction process often depends heavily on extensive engineering experience, making it difficult to achieve rapid methodological iteration and innovation. Some previous highlight aerial robot learning works are listed in Tab.\ref{tab:comparison}. The last column outlines the task design principles used in previous approaches which support our task-oriented framework. Comparing with these, our framework and workflow differ significantly from existing works that primarily focus on methodological innovation or simulator. Instead, we emphasize the evaluation and integration of proven or widely accepted approaches, provide a comprehensive explanation of key principles for task design, and highlight the challenges of policy transfer along with practical solutions. In conjunction with the aforementioned design principles and techniques, the simulator, software, firmware, and hardware modules we develop collectively form a complete and cohesive workflow. Based on this, our task-oriented training leverages DRL to train policies and enable zero-shot sim-to-real transfer to real-world aerial robots. \textbf{We hope this work serves as a stepping stone to inspire and enable more researchers to quickly engage in practical exploration within this field, rather than being hindered by limitations in infrastructure.}

Our research has also identified key limitations, especially in the sampling efficiency and generalization of DRL. Model-free RL methods often rely on vast amounts of trial-and-error interactions, making them computationally expensive and data-inefficient. Vectorized parallel simulation with simultaneous sampling in this work is one solution; however, it does not mask the inherent inefficiency of the sampling. Recently, differentiable simulations enable gradient-based optimization by backpropagating through environment dynamics, which can drastically reduce the number of required training samples\cite{Heeg2024,Zhang2024a}. Although it offers a promising direction for improving sampling efficiency compared to traditional model-free RL, its advantages are not yet fully realized in complex, real-world scenarios. The applicability of differentiable simulation is currently constrained by its reliance on fully differentiable environments—a requirement that many physical or stochastic systems do not satisfy. Bridging this gap could unlock more efficient training paradigms but remains an open research challenge. 

Generalization in task-oriented DRL policies remains a critical hurdle. While many DRL agents perform well on narrowly defined tasks, they often struggle when exposed to even slight variations in deployment scenarios. The most direct and effective approach to improving generalization is scaling up model parameters and training data. The success of large-scale models—such as large language models (LLMs) and vision-language models—demonstrates that scaling laws, where performance predictably improves with increased model size, data diversity, and computational resources, lead to remarkable generalization abilities. Analogously, aerial robot policies trained on massive, multi-domain datasets—combining simulation and real-world data across varying weather and obstacle conditions—are expected to exhibit stronger out-of-distribution (OOD) generalization. Embodied learning with large-scale models in aerial robotics thus offers a promising direction toward achieving robust generalization. Nonetheless, given the safety-critical and low-latency demands of aerial robot control, careful balance between model size and inference speed remains essential.

\section{Methods}

\subsection{Deep Reinforcement Learning}
We model the aerial robot learning problem into a Markov Decision Process (MDP) problem and solve it by DRL, which can be mathematically expressed as $\left \langle \mathcal{S}, \mathcal{A}, \mathcal{T}, \mathcal{S'}, R \right \rangle$, where $\mathcal{S}$ is the state space and $\mathcal{A}$ is the action space. $\mathcal{T}$ is the state transition function, and $\mathcal{T}\left( s'|s, a \right)$ represents the transition probability distribution of the system when transiting into the next state $s' \in \mathcal{S'}$ from state $s \in \mathcal{S}$ after taking action $a \in \mathcal{A}$. Besides, $R$ denotes the reward function, where $r = R\left(s, a \right)$ is the reward given to the agent after taking action $a$ under state $s$. Our aim is to find the optimal policy $\pi^*$, to maximize the expected value of cumulative discount reward:
\begin{equation}
\begin{aligned}
\pi^*=\mathrm{arg} \underset{\pi}{\mathrm{max}}J_{\pi}=\sum_{t=0}^{\infty}\gamma^{t}\cdot \mathbb{E}\left [ R(s_t, a_t)|\pi \right ]
\end{aligned}
\end{equation}
where $s_t\in \mathcal{S}$ and $a_t\in \mathcal{A}$ are the state and action of the agent at time step $t$, and $\gamma$ denotes the discount factor.

We use Proximal Policy Optimization (PPO) algorithm\cite{Schulman2017b} to train the quadrotor policy, with a discount factor $\gamma$ equal to $0.95$. We use a shared actor-critic backbone with learning rate $3\times 10^4$. Since our contribution focuses on the entire sim-to-real system rather than the algorithm itself, we use the default parameters from \texttt{rl\_games}\cite{Makoviichuk2021} without any specific tuning of the algorithm. The simulator's dynamics calculation time step $\bm dt$ is $0.01$ s. The image render frequency is slower than dynamics calculation, with a render timestep of $0.04$ s which is similar to the depth image in reality, because the real frequency of images varies from $25\sim30$ Hz. The experiments are completed on a 4090 GPU with 24G memory. During the sampling process, we encourage using a large batch size to improve network convergence, especially in DRL where the search space is extremely large. In this work, we do not set any learning curriculum but use hierarchical training in dynamics obstacle avoidance task.

\subsection{Observation \& Action Space}
We commonly use typical quadrotor states as proprioceptive observation, illustrated in Fig.\ref{fig:pipeline}, involving the position, velocity, attitude, and angular velocity of the quadrotor root, or the difference between its target. Rotation matrix is used to represent the attitude due to the quaternion representation of a rotation being not unique. Therefore, the basic dimension of proprioceptive observation is $18$. Under the ego-centric observation design mode, we use states under $(\bm x_{\mathcal{E}},\bm y_{\mathcal{E}},\bm z_{\mathcal{E}})$ frame. Moreover, the exteroceptive observation is derived from a depth image captured from the forward-facing view. This depth image is processed through a simple convolutional neural network (CNN), a ResNet18, or a pretrained auto-encoder to extract features, where the feature dimensions can be adjusted depending on the complexity of the environment. Then, the policy network accepts the concatenation of both proprioceptive and exteroceptive observation and generates the action. Thus, the action space depends on the selected control levels. The action dimension is $4$ in $PY$, $LV$, $CTBR$, and $SRT$ mode, but $5$ in the $CTA$ mode.

\subsection{Bridging Gaps in Sim-to-real}
\subsubsection{System Identification} 
The role of system identification in policy transfer is well-established and needs no further emphasis\cite{Kaufmann2022,Chen2024}. The design of the controller relies on an accurate description of the system model. System identification enables precise acquisition of the dynamic characteristics. This enhances the realism of simulation, ensuring that control policies designed are more effective when deployed in real-world scenarios. However, thanks to the differential flatness of quadrotor, it enables dynamic parameterization, meaning the RL policy in training only needs to address uncertainties within the flat output space instead of the full state space, greatly simplifying system identification\cite{Song2023}. Moreover, by leveraging large amount of sampling and domain randomization, DRL inherently addresses model uncertainty, diminishing the need for accurate system modeling. Here we emphasize three critical aspects that must be carefully addressed during the quadrotor system identification process: hovering throttle, air drag, and system delay. Hovering throttle refers to the precise throttle value required to maintain a stable hover which is necessary to counter the gravity, and will be discussed in details later. Air drag plays a significant role in the quadrotor's translational dynamics, especially at higher velocities\cite{Sun2022a,Kaufmann2023}. It introduces nonlinear effects that must be captured to ensure realistic modeling and robust control performance. Finally, the system delay\cite{Zhang2024a}, including sensor latency, actuator response time, and computation delay, affects the closed-loop behavior and must be quantified to build a reliable and predictive model of the entire system.

\subsubsection{Hover Throttle Learning} 
In low-speed and low-maneuverability flights, system failures due to air drag and system delay are uncommon. While the hover throttle parameter estimation is typically the primary consideration during system identification. In the implementation of PX4 autopilot, the hover throttle is also a crucial setting. It is usually expressed as a normalized value between 0 and 1. It determines how the controller allocates thrust to counteract gravity and maintain altitude stability during hover. This setting enhances compatibility with different aircraft, making it more general. Moreover, this setting could decouple the altitude and attitude control, making the quadrotor more flexible. To ensure that the policy trained in simulation can maintain similarly stable outputs in real-world scenarios while preserving accuracy in altitude control, we recommend designing the reward during training to gradually guide the policy's total throttle output to converge to the actual hover throttle value. Otherwise, an inaccurate hover throttle can result in steady-state errors along the $\bm z$ axis and even lead to control divergence in the altitude loop.

\subsubsection{Learning Continuity of Actions} The continuity of policy output is a key factor in sim-to-real control\cite{Chen2024}. In simulation, the dynamics computation time step $\bm dt$ is strictly consistent. However, the control frequency may fluctuate slightly depending on operating conditions in practice. The latency, which can impact the completeness of actuator execution, should be considered. For instance, in learning to maintain hovering, the agent might acquire behaviours involving rapid oscillations within the $(0,1)$ range, as shown in Fig.\ref{fig:gear}. While the altitude may remain well performed in the simulator, when transferred to a physical quadrotor, the motors cannot execute instantaneous and high-frequency accelerations and decelerations, leading to a crash. Therefore, a continuity reward for actions is essential and should be incorporated throughout the entire learning process.

\subsubsection{Reference Trajectory Play} 
As aforementioned, in tracking tasks, spatiotemporal awareness guidance is used for giving a reference in the near future. So that the policy could follow up the complicated trajectory. The previous work\cite{Chen2024} has shown that RL methods with appropriate guidance can achieve high-precision trajectory tracking when a given velocity is specified. These experiments lay on the ideal environments such that the velocity can maintain steady. However, the situation when quadrotor is affected by disturbances is not taken into consideration. In reality, quadrotors might be affected by external factors like wind and inaccurate pose estimation, which can prevent them from flying at the expected speed. The result can be found in Fig.\ref{fig:track-avoid}. If the quadrotor cannot keep up with the playback of the trajectory guidance, the tracking policy may fail or even collapse. Therefore, in tasks involving external spatiotemporal guidance, it is important to consider both appropriate reference trajectory replay and external disturbances.

\subsubsection{Domain Randomization (DR)}
Although various domain randomization methods are already common in sim-to-real applications, we want to emphasize the importance of their effectiveness here. The role of domain randomization is not only to narrow the gap between simulation and reality but also to encourage exploration through parameter and state randomness introduced by random initialization. This is particularly critical in DRL with image-based inputs; otherwise, the policy may easily get trapped in local optimal. In this work, we used VIO as the data for state estimation of the quadrotor, without any external ground truth observation inputs. Therefore, domain randomization for the quadrotor's fundamental states is particularly crucial. Without this, the agent may encounter unseen states, which could lead to a failure or collapse. To ensure the generalization of images during sim-to-real transfer, DR should also be applied to the visual information during simulation training\cite{Loquercio2021}. We process the depth maps generated in simulation by adding Gaussian noise and multiplicative noise to mimic sensor noise and errors. Additionally, we apply blurring to the depth to mimic lens distortion, motion blur, or inaccurate focus. Furthermore, we randomly scale and offset the depth to simulate sensor calibration discrepancies, changes in ambient lighting, or inconsistencies in depth measurement. 

\section*{Data availability}
All data necessary to evaluate the conclusions in the paper are available in the paper or the Supplementary Information.

\section*{Code availability}
Simulation codes, software and firmware for fight platform are publicly accessible in the GitHub repository: \url{https://github.com/emNavi/AirGym}, \url{https://github.com/emNavi/rlPx4Controller}, \url{https://github.com/emNavi/AirGym-Real}, \url{https://github.com/emNavi/control_for_gym}.



\section*{Acknowledgements}
We thank Wantong Qin for her assistance in the design and development of the flight platform, and Kai Tang for his support in VIO flight testing.

\section*{Author contributions}
K.H and H.W designed the full flight system, including the simulator and physical platform. K.H built the framework and provided the visualization. K.H and J.Y.C wrote the paper. K.H and Y.L provided reward design in RL. J.T.C and D.G provided testing resources and validation. H.L, X.J, and X.Z were responsible for the project supervision and conceptualization.

\section*{Competing interests}
The authors declare no competing interests.

\section*{Additional information}

\textbf{Correspondence} and requests for materials should be addressed to Huaping Liu.

\newpage
\appendix
\section{Dynamics Model}
The world $\mathcal{W}$ and body $\mathcal{B}$ frames with the orthonormal basis in this work are defined: 
$\{\bm{x}_\mathcal{W},\bm{y}_\mathcal{W},\bm{z}_\mathcal{W}\} $, $\{\bm{x}_\mathcal{B},\bm{y}_\mathcal{B},\bm{z}_\mathcal{B}\} $. The frame $\mathcal{B}$ is at the root of the quadrotor. Symbols with subscript mean variables under the corresponding coordination frame. The system dynamics can be expressed by the following equations:
\begin{equation}
\begin{bmatrix}
\dot{\bm{p}}_{\mathcal{W}} \\
\dot{\bm{q}}_{\mathcal{B}} \\
\dot{\bm{v}}_{\mathcal{W}} \\
\dot{\bm \omega}_{\mathcal{B}} \\
\dot{\bm \Omega}
\end{bmatrix}
=
\begin{bmatrix}
\bm{v}_{\mathcal{W}} \\
\bm q_{\mathcal{B}} \cdot 
\begin{bmatrix}
0 & \bm \omega_{\mathcal{B}} / 2
\end{bmatrix}^\top \\
\frac{1}{m} \left( \bm q_{\mathcal{B}} \odot \bm f \right) + \bm g_{\mathcal{W}} \\
\bm J^{-1} \left( \bm \tau - \bm \omega_{\mathcal{B}} \times \bm J \bm \omega_{\mathcal{B}} \right) \\
T_\mathrm{m} \left( \bm \Omega_{\text{cmd}} - \bm \Omega \right)
\end{bmatrix}
\end{equation}
where $\bm p, \bm q, \bm v, \bm \omega, \bm \Omega$ represent position, quaterion, velocity, body rate (angular velocity), propeller rotation speed, respectively. $\bm \Omega_{\mathrm{cmd}}$ is the command of rotational speed, and $T_\mathrm{m}$ is the motor time constant, which can be estimated by rotational speed testing. $\bm J=\mathrm{diag} (I_x, I_y, I_z)$ denotes diagonal moment of inertia matrix. $m$ is the mass of quadrotor. Note the quaternion-vector product is denoted by $\odot$ representing a
rotation of the vector by the quaternion as in $\bm q \odot \mathrm{\bm v} = \bm q \mathrm{\bm v} \bm q^-$,
where $\bm q^-$ is the quaternion’s conjugate. $\bm g_\mathcal{W} = [ 0, 0, -9.81 ]^\top$ denotes gravity. $\bm f$ and $\bm \tau$ are the collective force and the torque produced
by all propellers. The quantities are calculated as follows:
\begin{equation}
    \bm f=\sum_i f_i, \bm \tau=\sum_i \tau_i+\bm r_{P, i} \times f_i
    \label{eq:thrust and torque}
\end{equation}
where $\bm r_{P,i}$ denotes the location of propeller $i$ expressed in the body frame, $f_i$ and $\tau_i$ are forces and torques generated by the $i$th propeller, where we have:
\begin{equation}
{f}_i( \Omega) = \begin{bmatrix} 0 & 0 & c_l \cdot  \Omega^2 \end{bmatrix}^\top
,
{\tau}_i( \Omega) = \begin{bmatrix} 0 & 0 & c_d \cdot  \Omega^2 \end{bmatrix}^\top
\label{eq:coeffients}
\end{equation}
where $c_l$ and $c_d$ are corresponding thrust and drag coefficients that can be obtained by system identification.
\begin{figure}[h]
    \centering
    \includegraphics[width=.8\linewidth]{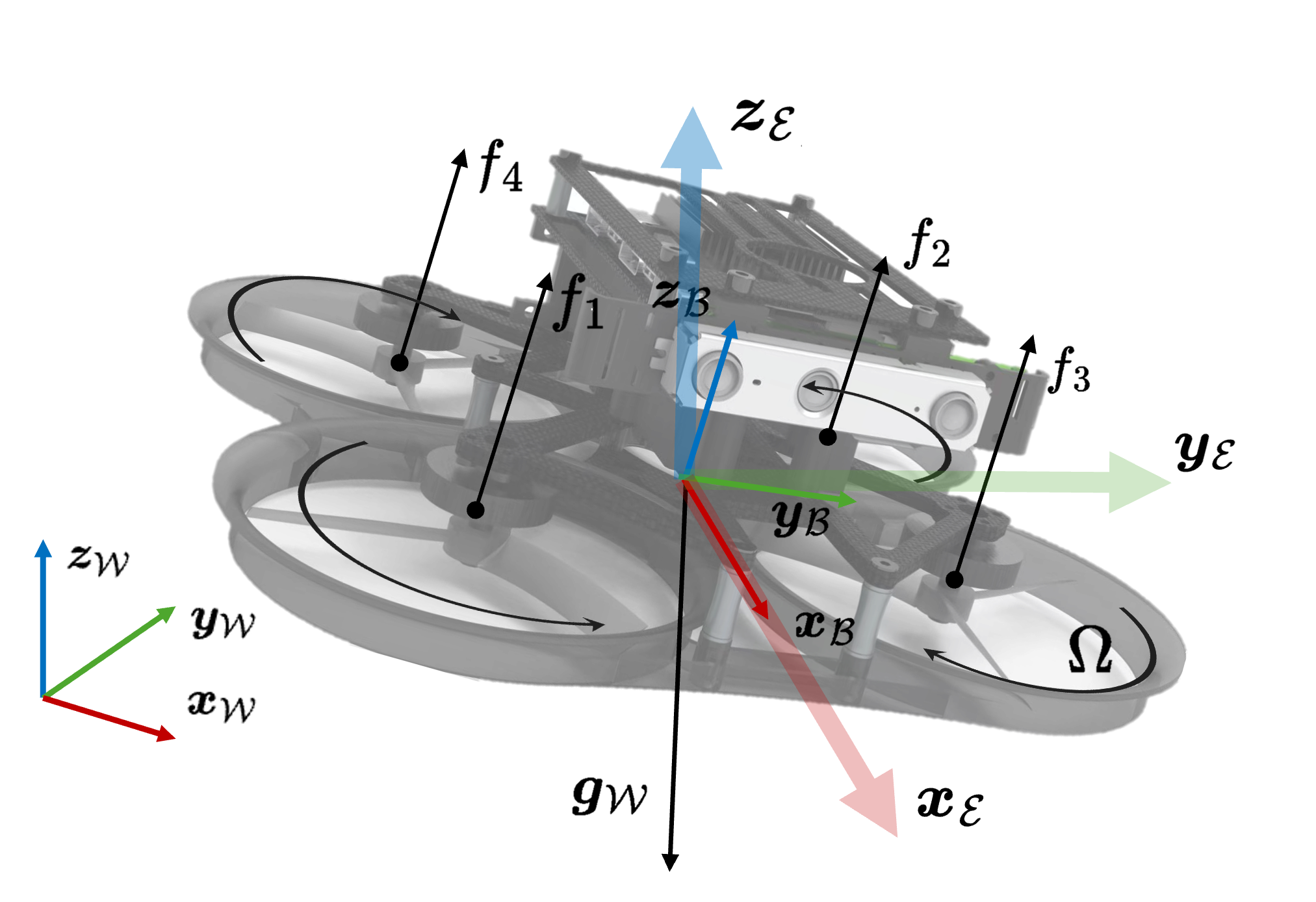}
    \caption{Diagram of the \texttt{X152b} depicting the world and body frames and
    propeller numbering. The background image showcases the construction and visualization of the \texttt{X152b} quadrotor in IsaacGym.}
    \label{fig:dynamics}
\end{figure}

\section{Task Design \& Reward Settings}
\subsection{hovering} Hovering is the most fundamental task of RL-trained quadrotor. We defined an 18 dimensional quadrotor state: 
\[(\mathrm{R}^{\rightharpoonup},p_x,p_y,p_z,v_x,v_y,v_z,\omega_x,\omega_y,\omega_z)\]
where $\mathrm{R}^{\rightharpoonup}$ denotes the flattening of rotation matrix $\mathrm{R}$. Then the differential between the current state and expected state is used as the observation. It brings advantages that all states will converge to zero, leading to a more uniform gradient in the data probability distribution. Additionally, by modifying the desired state during flight, commands can be sent to the quadrotor, enabling waypoint control similar to that in traditional flight controllers. The reward function is set as below:
\begin{align}
    \bm R &= R_{smooth}+R_{effort}+R_{pos}+R_{throttle}\\
    &+R_{pos}\times(R_{ups}+R_{spin}+R_{heading}+R_{vel\_dir})\\
    R_{smooth} &= K_{1} \times e^{- \|\Delta a \|}\\
    R_{effort} &= K_{2} \times \sum_i^n (1-\Omega_i), n=4\\
    R_{pos} &= K_3 / (1 + K_4 \times\| \Delta \bm
    p \|^2) \\
    R_{throttle} &= K_5 \times (1-|T_{hover}-T|)\\
    R_{ups} &= K_6 \times (\bm q \mathrm{\bm v}_z \bm q^{-}+1)^2 \\
    R_{spin} &= K_7 / (1+\omega_z^2)\\
    R_{heading} &= K_8 / (1+(\Delta \psi)^2)\\
    R_{vel\_dir} &= K_9 \times e^{- \bm v \cdot \bm d /\pi}
\end{align}
where $R_{smooth}$ denotes the reward for continuous actions; $R_{effort}$ is to encourage quadrotor to consume as little energy as possible; $R_{pos}$ is position reward related to absolute position error; $R_{throttle}$ helps quadrotor to learn the expected hover throttle $T_{hover}$; $R_{ups}$ maintains a stable horizontal attitude and avoids excessively large roll and pitch angles; $R_{heading}$ guides the target yaw direction $\psi$ and $R_{spin}$ tries to fix the spin around $\bm z^{\mathcal{B}}$-axis; finally, $R_{vel\_dir}$ leads to a straight flying direction from the current point to the target, using the dot product between the current flight direction vector $\bm v$ and the vector pointing to the target $\bm d$. Note that the $R_{throttle}$ equals zero when using $PY$ and $LV$ control modes that do not learn a collective thrust.

\subsection{tracking} The fundamental observation in this task is insufficient. A common approach is to use trajectory playing, adding a sequence of the coming trajectory. Herein, we set an observation sequence of length $30$ to store the coordinates of the upcoming $10$ points, guiding quadrotor to follow the trajectory. Therefore, we do not define the observation by the differential, while we extend the observation dimension to 48: $(\mathrm{R}^{\rightharpoonup},p_x,p_y,p_z,v_x,v_y,v_z,\omega_x,\omega_y,\omega_z,\mathrm{ref}^{\rightharpoonup})$, where $\mathrm{ref}^{\rightharpoonup}$ is a concatenation of the positions of ten reference points in the future. The reference trajectory is calculated by the parametric equations of a lemniscate (figure-eight curve), which has the longest side to range from $(-6, 6)$ meters:
\begin{equation}
    \bm x(t) = \frac{3 \sin(kt)}{1 + \cos^2(kt)},\;
    \bm y(t) = \frac{3 \sin(kt) \cos(kt)}{1 + \cos^2(kt)},\;
    \bm z(t) = 1
\end{equation}
where $k$ denotes a scaling factor to control the speed of quadrotor, directly determining the duration to complete one lemniscate lap. We also add distance reward part in this reward setting. Reward functions:
\begin{align}
    \bm R &= R_{smooth}+R_{effort}+R_{dist}+R_{throttle}\\
    &+R_{dist}\times(R_{ups}+R_{spin}+R_{heading})\\
    R_{smooth} &= K_{1} \times e^{- \|\Delta a \|}\\
    R_{effort} &= K_{2} \times \sum_i^n (1-\Omega_i), n=4\\
    R_{dist} &= K_3 / (1+ K_4 \times \| \Delta \bm p \|^2)\\
    R_{throttle} &= K_5 \times (1-|T_{hover}-T|)\\
    R_{ups} &= K_6 \times (\bm q \mathrm{\bm v}_z \bm q^{-}+1)^2 \\
    R_{spin} &= K_7 / (1+\omega_z^2)\\
    R_{heading} &= K_8 / (1+(\Delta \psi)^2)
\end{align}
We have similar reward items compared with task \texttt{hovering}. $R_{dist}$ is used to replace the position error item $R_{pos}$ in \texttt{hovering}. In $R_{dist}$, $\bm p$ is the distance between the current position and the expected position in the trajectory sequence.

\subsection{target hitting} This task has the same state description with \texttt{hovering}. The state of balloon is randomly generated at every begin of the episode. Two kinds of rewards are designed. Dense reward encourages quadrotor to learn how to fly and move to the balloon, then a large sparse reward is given if the balloon is hit. Reward functions:
\begin{align}
    \bm R &= R_{smooth}+R_{effort}+R_{guidance}+R_{ups}\\
    &+R_{heading}+R_{hit}\\
    R_{smooth} &= K_{1} \times e^{- \|\Delta a \|}\\
    R_{effort} &= K_2 \times e^{-\sum_i^4 \|a\|^2}\\
    R_{guidance} &= K_3 \times (\|\bm p^b-\bm p_{t-1}\| - \|\bm p^{b}-\bm p_t\|)\\
    R_{ups} &= K_4 \times (\bm q \mathrm{\bm v}_z \bm q^{-}+1)^2 \\
    R_{heading} &= K_5 / (1+(\Delta \psi)^2)\\
    R_{hit} &= K_6 \; \mathrm{if \; balloon \; is \; hit}, \; \mathrm{else} \; 0.
\end{align}
where $R_{guidance}$ is a guidance reward that encourage quadrotor move from the start point $\bm p_s$ to the target balloon $\bm p^{b}$. A big positive sparse reward $R_{hit}$ is added once the balloon is hit. Also, a great hitting brings the termination of an episode.

\subsection{avoidance} Avoiding a dynamic obstacle by vision is always a exciting task for autonomous robots. We define the quadrotor state by ego-centric observation design: $(d_x,d_y,d_z,e_x,e_y,e_z,v_x,v_y,v_z,\omega_x,\omega_y,\omega_z,a)^{\mathcal{E}}$, where $d_{*}$ denotes the goal direction, $e_{*}$ denotes the Euler angle. All states are under the ego coordinate frame $\mathcal{E}$. We use $CTBR$ control mode which means the action $a$ has 4 dimensions. Besides, a simple CNN($[1,16,5,2,2]>[16,32,3,2,1]>[32,64,3,2,1]>['AdaptiveAvgPool',1]$) is set as a feature encoder to extract depth feature from binocular vision. At the beginning of an episode, we generate a cube with a suitable throwing speed and throwing angle. The onboard camera outputs the depth arrays with $(1, 212, 120)$ as the input of feature extractor. The feature extractor infers a 30-dimensional feature vector and makes a concatenation with other states for final policy MLP. Reward functions:
\begin{align}
    \bm R &= R_{smooth}+R_{effort}+R_{throttle}+R_{pose}\\
    &+R_{alive}+R_{pose}\times (R_{ups}+R_{spin})\\
    R_{smooth} &= K_{1} \times e^{- \|\Delta a \|}\\
    R_{effort} &= K_2 \times e^{-\sum_i^4 \|a\|^2}\\
    R_{throttle} &= K_3 \times (1-|T_{hover}-T|)\\
    R_{pose} &= K_4 / (1+\|\Delta [\bm p,\theta,\phi,\psi]\|^2)\\
    R_{ups} &= K_5 \times (\bm q \mathrm{\bm v}_z \bm q^{-}+1)^2 \\
    R_{spin} &= K_6 / (1+\omega_z^2)\\
    R_{alive} &= K_7 \; \mathrm{if \; alive, \; else \;} K_8
\end{align}
Here we use item $R_{pose}$ to indicate both position and orientation, $\theta,\phi,\psi$ represent pitch, roll, yaw angles respectively. Besides, alive reward is also used in this task, while it is a dense reward: if the quadrotor stay alive for every step, it obtains a small positive reward; if the quadrotor is hit by the throwing object, it receives a big negative penalty.

\subsection{planning} this work utilize the same observation definition in \texttt{avoidance}. However, navigation in unknown environments is much more complex than any other tasks. As the most important trick in \texttt{planning}, we introduce Euclidean signed distance field ($ESDF$) to describe the distance from quadrotor to the closest obstacle, and propose an $ESDF$ reward to guide the quadrotor to the target, shown in the sub-figure (b) of the pipeline illustration. There is a problem: how to establish the connection between depth arrays $D\in\mathbb{R}^{H\times W}$ and $ESDF$ distance $x_{esdf}$? In \texttt{planning}, we directly use the value of the point with the smallest depth in the depth array to approximate the $ESDF$ distance: \[x_{esdf}=D_{\mathrm{min}}=\min_{i,j}D_{i,j} \]
Rely on this, we define the $ESDF$ reward as $R=1-e^{-x^2}$, to encourage quadrotor to pull itself away from the nearest $ESDF$ point. Reward functions:
\begin{align}
    \bm R &= R_{smooth}+R_{effort}+R_{throttle}+R_{guidance}\\
    &+R_{speed}+R_{height}+R_{heading}+R_{ups}\\
    &+R_{guidance} \times (R_{esdf}+R_{alive})+R_{goal}\\
    R_{smooth} &= K_1 \times (\| \Delta a \|+\| \bm \omega_{\mathcal{E}}\|)\\
    R_{effort} &= K_2 \times e^{-\sum_i^4 \|a\|^2}\\
    R_{throttle} &= K_3 \times (1-|T_{hover}-T|)\\
    R_{guidance} &= K_4 \times (\|\bm p^{target}-\bm p_{t-1}\|-\| \bm p^{target}-\bm p_t\|)\\
    R_{speed} &= -K_5 \times (1-e^{-K_6\times (\|\bm v^\mathcal{E}_x\|-K_7)^2})\\
    R_{height} &= \min(\min(p_z-K_8,0),K_9-p_z)\\
    R_{esdf} &= K_10 \times (1- e^{-K_11 \times x_{esdf}^2})\\
    R_{ups} &= K_{12} \times (\bm q \mathrm{\bm v}_z \bm q^{-}+1)^2 \\
    R_{alive} &= K_{13} \; \mathrm{if \;} x_{esdf} > K_{14}\; \mathrm{else} \; 0\\
    R_{goal} &= K_{15} \; \mathrm{if \;} \mathrm{goal \; reached,}\; \mathrm{else} \; 0
\end{align}

In this task, most of observations are under $\mathcal{E}$ frame. $K_7$ denotes the maximum expected speed during the flight. $K_8,K_9$ are the minimum and maximum expected flight height. We use the minimum value in the depth array to replace $ESDF$ distance $x_{esdf}$. Furthermore, we also use $ESDF$ distance to judge whether the quadrotor is crashed or not by compute if $x_{esdf}$ less than range $K_{14}$.

\section{Control Results under Different Modes}
\begin{figure*}
    \centering
    \subfigure[\texttt{Hovering} training reward]{\includegraphics[width=0.245\textwidth]{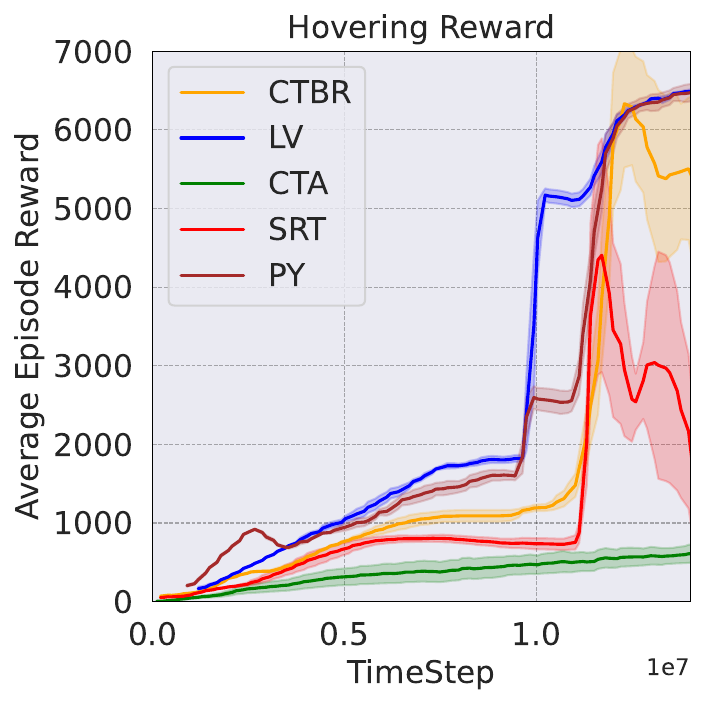}}
    \subfigure[\texttt{Target Hitting} training reward]{\includegraphics[width=0.245\textwidth]{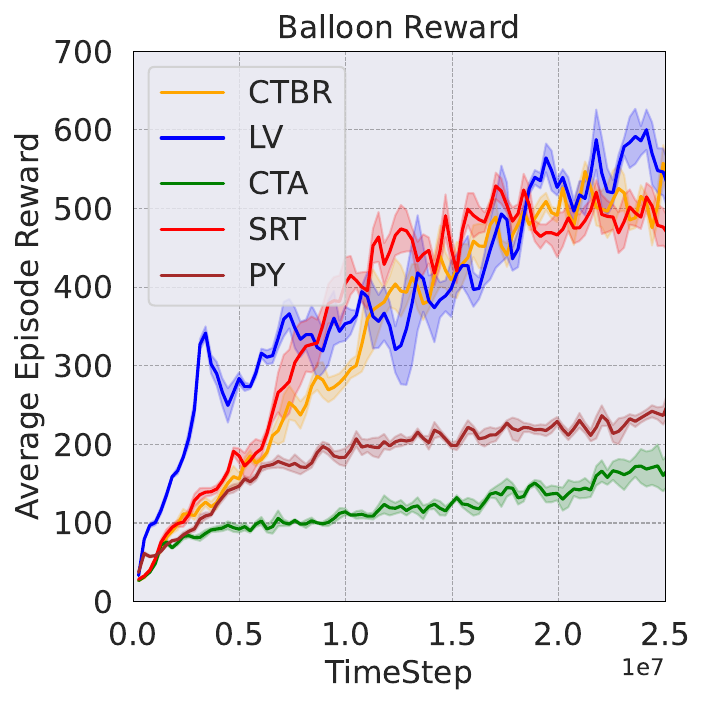}}
    \subfigure[\texttt{Tracking} training reward]{\includegraphics[width=0.245\textwidth]{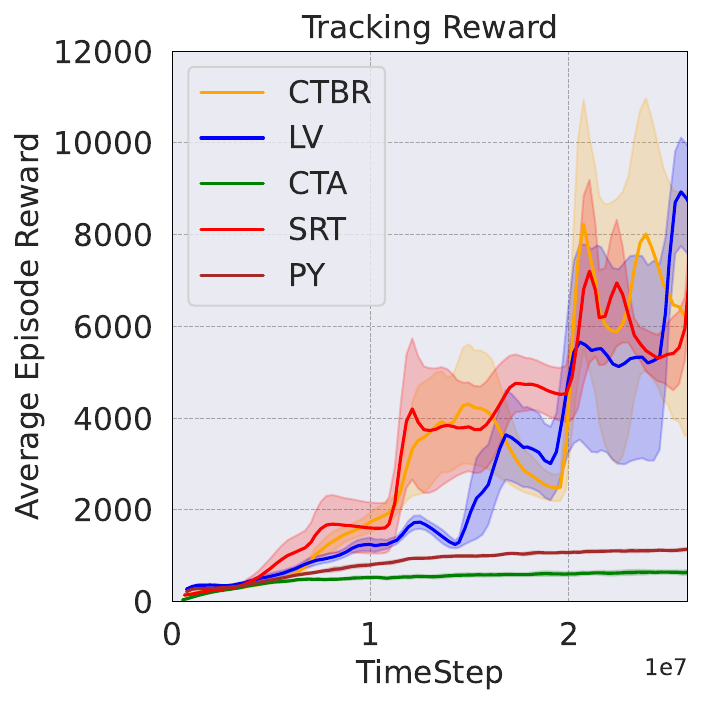}}
    \subfigure[\texttt{Planning} training reward]{\includegraphics[width=0.245\textwidth]{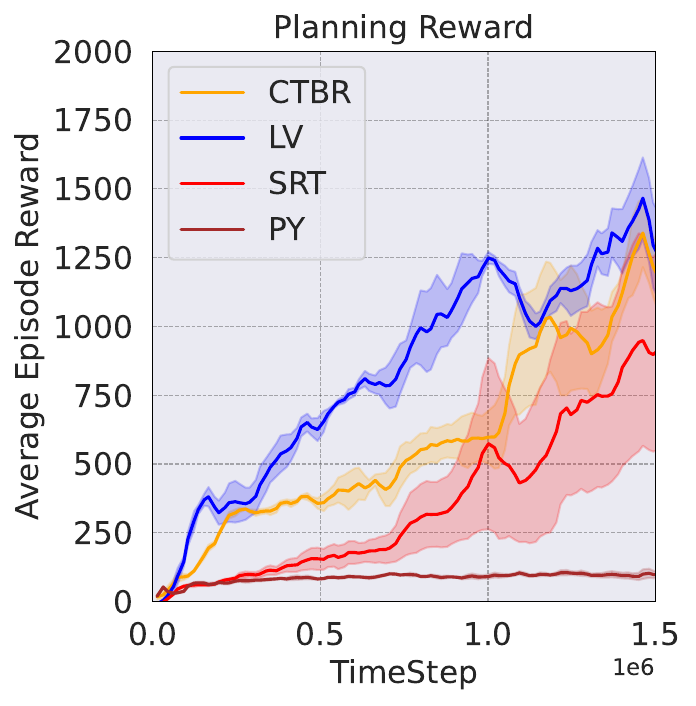}}
    \\
    \centering
    \subfigure[\texttt{Hovering} errors convergency]{\includegraphics[width=0.245\textwidth]{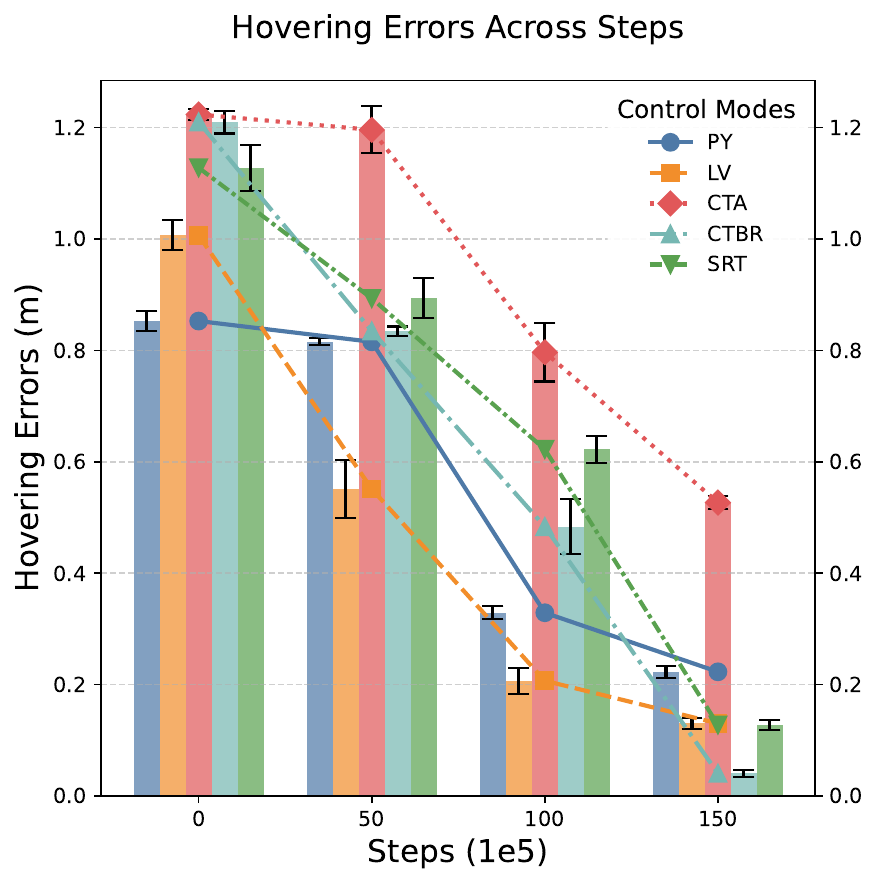}}
    \subfigure[\texttt{Target Hitting} SR convergency]{\includegraphics[width=0.245\textwidth]{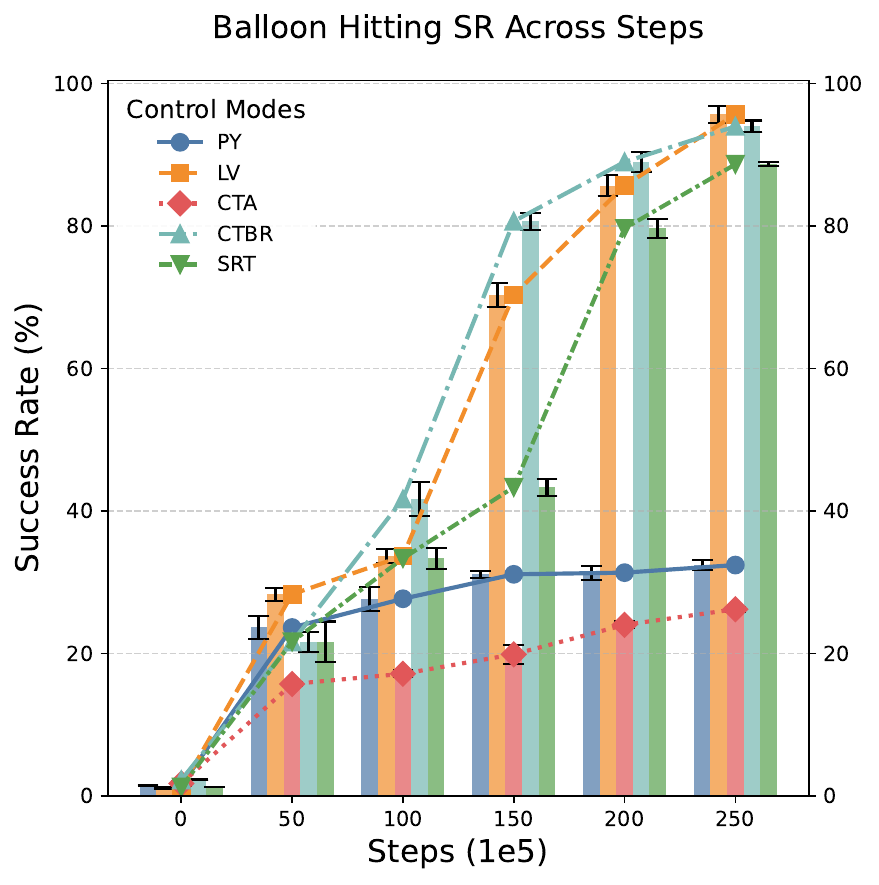}}
    \subfigure[\texttt{Tracking} errors convergency]{\includegraphics[width=0.245\textwidth]{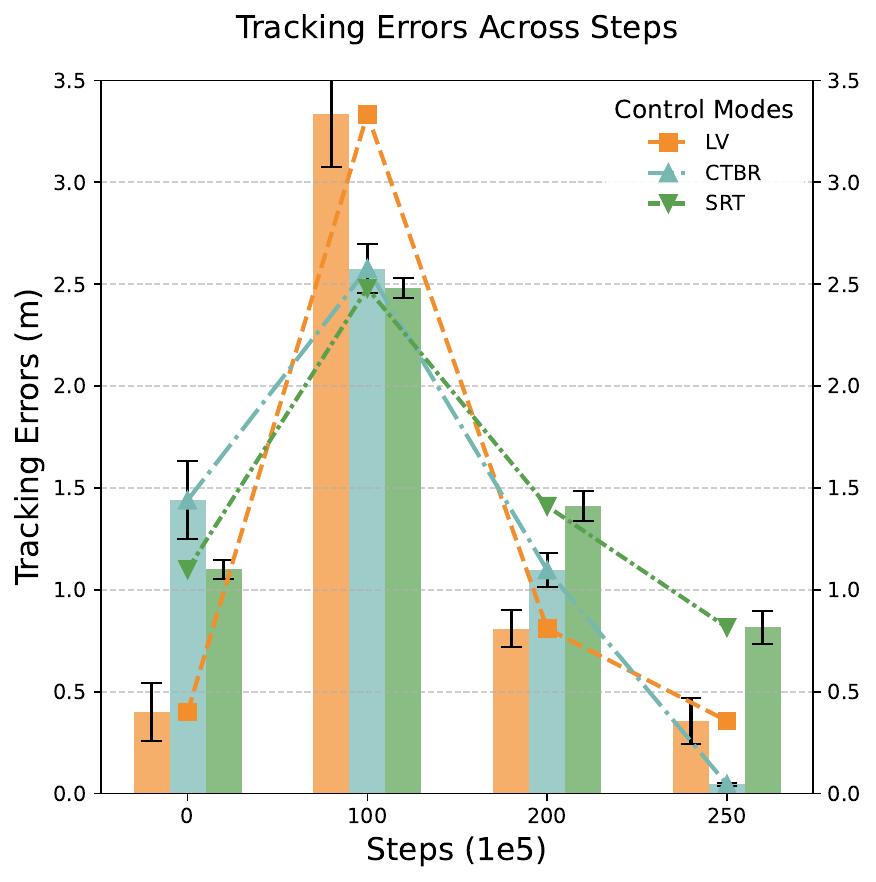}}
    \subfigure[\texttt{Planning} SR convergency]{\includegraphics[width=0.245\textwidth]{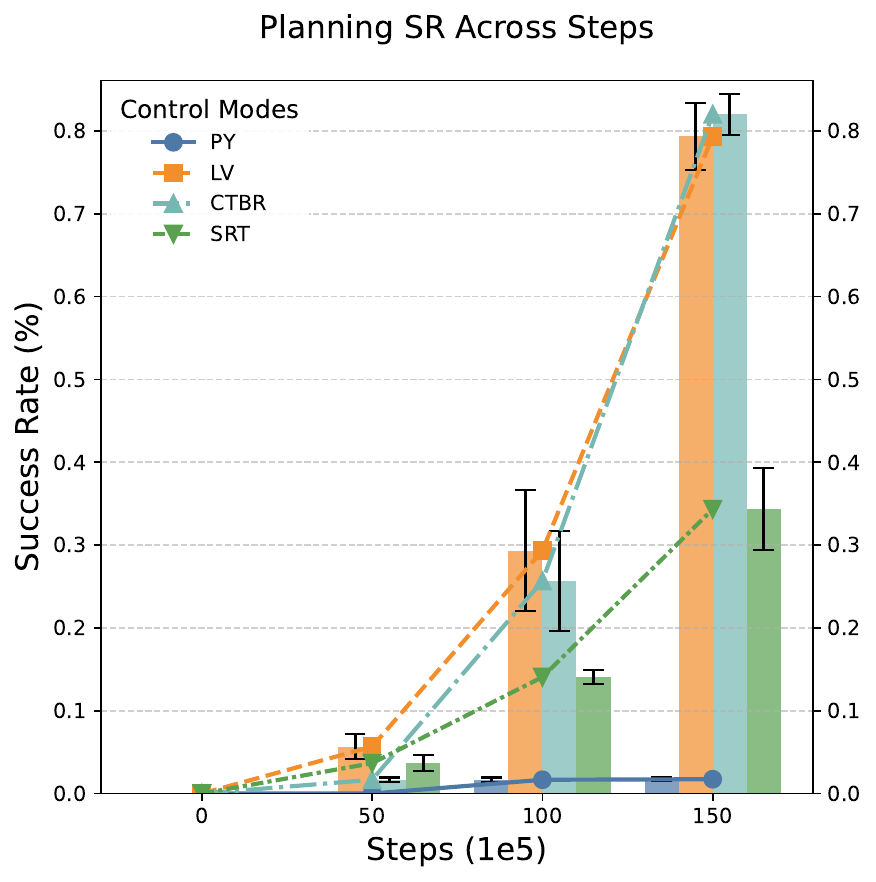}}
    \caption{Training indexes changing in four tasks: \texttt{hovering}, \texttt{target hitting}, \texttt{tracking}, and \texttt{planning}. Task \texttt{avoidance} is not illustrated because it need training curriculum and the reward makes no sense to refer. (a-d) present the reward changes during the training process. (e) and (g) show the errors in task \texttt{hovering} and \texttt{tracking}; (f) and (h) are success rate of task \texttt{target hitting} and \texttt{planning} respectively. }
    \label{fig:reward}
\end{figure*}
\begin{figure*}[t]
    \centering
    \subfigure[Control Pipeline for Different Modes]
    {\includegraphics[width=0.72\textwidth]
    {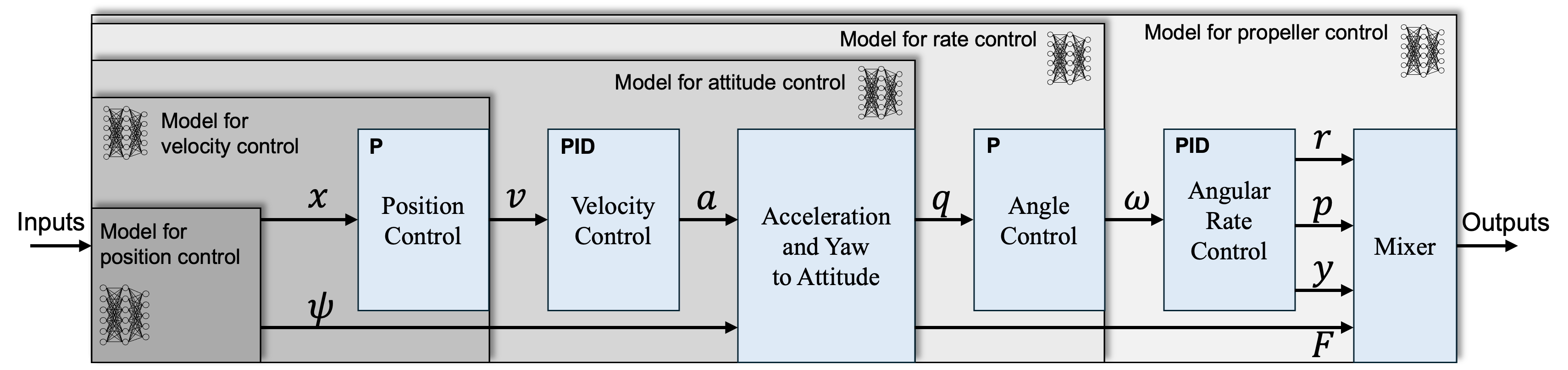}}
    \subfigure[Hovering Error]{\includegraphics[width=0.17\textwidth]{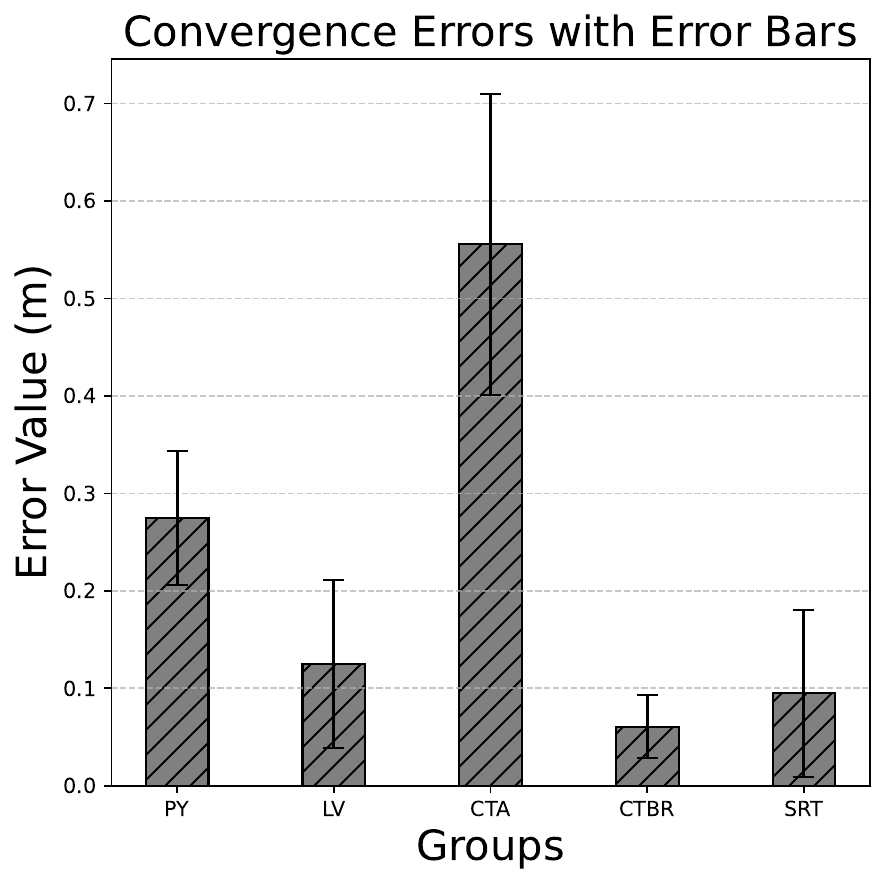}}
    \caption{Control pipeline under different control modes. Five modes are validated: position ($PY$), velocity ($LV$), angle ($CTA$), angular rate ($CTBR$), propeller ($SRT$). Errors after converging are shown in (b).}
    \label{fig:ctl pipeline}
\end{figure*}

First, we validate the performance of tasks in the \texttt{AirGym} using \texttt{rlPx4Controller}'s hierarchical controllers combined with NN-based policy control, illustrated in Fig.\ref{fig:ctl pipeline}. Our experiments involve five control levels: $PY$, $LV$, $CTA$, $CTBR$, and $SRT$. The RL-trained policy in each level of control mode generates the corresponding expected command, then \texttt{rlPx4Controller} executes the remaining loops of control. The training rewards are shown in Fig.\ref{fig:reward}(a). From the point of training speed, the more controller layers replaced by the RL policy, the slower the learning process becomes. This trend aligns well with intuition, as the neural network requires more time to learn the attitude stabilization of the quadrotor's inner loop. The slow training under $CTA$ mode is a special case. This may mean the attitude loop commands are not as direct as $CTBR$ and $SRT$ when policies are applied to the quadrotor. These two control styles are closer to the actuators, allowing the network to directly learn their relationships within the dynamics system without dealing with intermediate complex control logic. In contrast, in $PY$ and $LV$, RL focuses on generating high-level commands for the outer loop, while the quadrotor itself can inherently maintain attitude stability, making it easier and faster to learn. Comparing the converging and hovering accuracy across these five modes, $CTBR$ demonstrates an impressive advantage. We conduct eight trials for each type of experiment and calculate the average of hovering errors, as shown in Fig.\ref{fig:reward}(e). In the scenario where the quadrotor is randomly initialized within a cubic space with a side length of $2$ m and ultimately converges to the point $(0, 0, 1)$, the performance of $CTBR$ can maintain hover accuracy within $0.1$ m. Other methods can also maintain stability, but they fail to achieve the same level of control accuracy and robustness as $CTBR$. This result verifies the conclusion in again that $CTBR$ shows a good performance and a strong resilience against dynamics mismatch, on different types of hardware.

In other tasks, we showcase the training performances of \texttt{target hitting} and \texttt{tracking}, which are shown in Fig.\ref{fig:reward}(b)(c). In \texttt{target hitting}, we find $LV$, $CTBR$, and $SRT$ control modes reach convergence much faster. Although rewards under $LV$ and $SRT$ obtain similar scores, the actual effects are not consistent, shown in Fig.\ref{fig:reward}(f). Comparing the success rate (SR) of these five modes, control under $CTBR$ and $LV$ obtain higher SR. In $LV$ control mode, the reward increases rapidly because the attitude loop in this mode is inherently stable, allowing the agent to focus only on learning how to move toward the target. However, the quadrotor under $SRT$ mode does not first learn to maintain inner-loop stability but instead directly learns how to crash into the target without considering its own attitude, which simplifies the task. In brief, the quadrotor does not need to learn how to fly while only needing to learn how to hit the balloon. Task \texttt{tracking} achieves satisfactory results in three control modes: $CTBR$, $LV$, $SRT$, illustrated in Fig.\ref{fig:reward}(c)(g) within limited timesteps. Here we record the tracking errors of these three control modes during the training process (tracking errors under $PY$ and $CTA$ are not described because its low SR and the value is meaningless). At the beginning, it is obvious that all control modes obtain errors less than $1.5$ meters since the quadrotor is always reset at the first expected point, especially under $LV$ modes because these two modes provide a robust inner-loop stability. After the training, $CTBR$ and $LV$ report a better tracking precision. Task \texttt{planning} is the most challenging one among all tasks since it faces much more uncertain circumstances. From the SR record in Fig.\ref{fig:reward}(h), $PY$ mode demonstrates obvious limitations for navigation purpose. $LV$ and $CTBR$ based learning shows an impressive advantage, with approaching $80\%$ SR of navigation. Considering form the standpoint of visual effect, we believe the $CTBR$ mode to deliver superior performance, as this mode enables direct attitude-loop control of the quadrotor, resulting in more agile flight characteristics.

In summary, control under $CTA$ mode shows limited performance under all tasks, with longer duration and lower reward. The reasons can be diverse. The most important one is we employ quaternions as the network output, but the property of conjugate equivalence in quaternions may result in multiple valid output representations for the same orientation. This can lead to an imbalanced distribution in the training data.

\section{Infrastructures}
\subsection{Quadrotor Hardware Device: X152b}
We design a quadrotor \texttt{X152b} based on drone racing frame by OddityRC, illustrated in Fig.\ref{fig:dynamics}. The diagonal distance of this quadrotor is $152 \mathrm{mm}$. It is equipped with the Rockchip RK3588s as onboard processor and a mini open-sourced PX4 autopilot NxtPX4. The onboard computer delivers over $6$ TOPs of AI performance, enabling us to perform lightweight neural network inference almost in real time directly onboard. Additionally, it integrates an Intel RealSense D430 camera for perception acquisition. To enhance the agility and flight endurance, the weight is strictly kept under $400 \mathrm{g}$ (excluding the battery). We report a more than $35 \mathrm{m\cdot{s^{-1}}}$ maximum flying speed under first personal view (FPV) mode and over $11 \mathrm{m\cdot{s^{-1}}}$ speed under sim-to-real mode. 

\subsection{Vectorized Simulation Platform: AirGym}
We develop \texttt{AirGym} platform based on IsaacGym and inspired by previous work Aerial Gym. Compared with it, we rebuild the flight dynamics and make massive changes to align reality and simulation. (\romannumeral1) We use the physical dynamics parameters and real robot mesh file to replace general simplified quadrotor in Aerial Gym. (\romannumeral2) We provide a new asset registor and manager, making code more concise and allowing it to load various classes and quantities of objects. Also, we add more assets to make the task more realistic. (\romannumeral3) We provide multiple levels of RL policies to control a quadrotor, including $PY$, $LV$, $CTA$, $CTBR$, and single-rotor thrust ($SRT$), by replacing the corresponding control loop in a cascade PID.

We realize dynamics by directly applying force and torque to a rigid body to compute the collective thrust $\bm f$ and torque $\bm \tau$ applied to the root of the quadrotor, as shown in Eq.\ref{eq:thrust and torque}. The relationship between propeller rotation speed $\Omega$ and both thrust $f$ and torque $\tau$ are nonlinear in reality, typically exhibiting a quadratic correlation in Eq.\ref{eq:coeffients}. Determining this function requires system identification. However, in controllers based on differential flatness (include PX4), there is a normalized hover throttle to counteract gravity, and we linearize the quadratic curve at this point because the variation in its second derivative is negligible. Therefore, in this work, we approximate the relationship between propeller speed and thrust as linear and use actual flight data to estimate the true hover throttle.

\subsection{Parallelized Geometric Controller: rlPx4Controller}
Implementing quadrotor flight in large-scale vectorized environments requires parallel flight control. To provide parallelized computing, one solution is to inherit the neural network (NN) forward propagation module in \texttt{PyTorch} to tensorfy vectors, such as Aerial Gym and OmniDrones. Nevertheless, current works have not fully aligned with the PX4 logic and parameters, which is critical in the engineering implementation of PX4. To address this issue, we implemented a parallel geometric controller strictly similar to PX4. This allows us to guarantee that the data trained in simulation aligns perfectly with the real-world system characteristics during the sim-to-real process, especially in holding the altitude and handling the hover throttle.

We rewrite the PX4 flight control in \texttt{C++} according to nonlinear quadrocopter attitude control, and enable parallel computation. While achieving the same mathematical calculation, a stringent correspondence between the simulation controller and PX4 is established.

Specifically, we extract and encapsulate the control part of the PX4 autopilot by pybind11, providing \texttt{Python} interfaces that can be directly invoked at various control levels. We give a simple example of the usage of \texttt{rlPx4Controller}. The code is accessible online, offering interfaces that are straightforward to install and easy to use:
\begin{lstlisting}
git clone \
  git@github.com:emNavi/rlPx4Controller.git
cd rlPx4Controller
pip install -e .
\end{lstlisting}

Using the \texttt{rlPx4Controller} is straightforward. All we need to do is import the \texttt{rlPx4Controller}, then instantiate a controller of the expected type. It can be a position ($PY$), velocity ($LV$), attitude ($CTA$), or rate ($CTBR$) controller, and then call its \texttt{update} function:
\begin{lstlisting}
from rlPx4Controller import ParallelRateControl, ParallelVelControl, ParallelAttiControl, ParallelPosControl
# According to which control layer is selected
# Use Rate control as an example
controller = ParallelRateControl(num_envs)
controller.set_status(status, dt)
cmd_thrusts = controller.update(actions)
\end{lstlisting}
\texttt{num\_envs} is the number of parallel computing processes. The function \texttt{set\_status} takes as input the status of the drone, including position, attitude in quaternion, linear velocity, and angular velocity, and settings for the controller's simulation interval \texttt{dt}. The function \texttt{update} takes actions as input and outputs the final command of thrusts. More detailed features and usages of \texttt{rlPx4Controller} can be found in the online documentation\footnote{\href{https://rlpx4controller.readthedocs.io/en/latest/index.html}{https://rlpx4controller.readthedocs.io/en/latest/index.html}}.

\subsection{Onboard ROS Inference Node: AirGym-Real}
As aforementioned, \texttt{AirGym} provides a platform for fast training in simulator. During Sim-to-Real, we need software similar to AirGym that can directly load the trained neural network on the onboard system, interact with sensor and flight control data, and perform real-time inference from states to actions. Particularly in the wild, where there is no direct pose information available, such as from motion capture systems, we can only rely on other tricky methods to obtain the required observation for the inference.

We implement the above functionality as a ROS node to obtain state from PX4 autopilot and camera, as well as to conduct the inference. To obtain full states of quadrotor in the wild, we run a visual-inertial odometry (VIO) onboard. VINS-fusion is selected for this task, which is achieved by extracting feature points from the environment captured by the camera and combining them with the IMU information from PX4 Autopilot. It is worth mentioning that PX4 firmware will estimate IMU bias using input position data by default, which could contaminate the raw IMU data and lead to a nonconvergent solution during the VIO fusion. To address this issue, we modify the PX4 firmware by removing the IMU bias estimation component, which is commonly used in engineering but is less crucial for research. Meanwhile, we ensure the position estimation in the EKF to ultimately approach the VIO output position, to avoid re-fusion of IMU data. Furthermore, it brings another convenience: 
the coordinate alignment between the VIO system and the PX4 controller is not a concern in our system. The effectiveness and robustness of VINS classes have been extensively validated in previous studies. Therefore, this paper will not elaborate on their performance in detail. The raw depth data undergoes preprocessing, including resizing and depth truncation, before being used to construct the observation input for neural network inference. Finally, the computed action is published to the ROS topic.

\subsection{PX4 Control Bridge: control\_for\_gym}
\texttt{control\_for\_gym} is a PX4 middleware stack to publish NN-inferenced commands to expected levels of PX4 control loop. Two functions are primarily implemented: constructing a finite state machine (FSM) to enable switching between trained policy control and classical PID control; and forwarding control commands to the PX4 Autopilot controller based on the selected control hierarchy.

During DRL training, quadrotors are often initialized from a specific state, such as hovering stably at a certain position. This requires the quadrotor to stabilize to the desired state before inference in reality, ensuring continuous control and improving the success rate of sim-to-real transfer. To this end, a simple FSM is constructed for providing steady switching, which is shown in the pipeline in the manuscript. While transferring policy to the wild, we first conduct an automatic taking-off behaviour and then hover in place to ensure the quadrotor starts to infer with an appropriate state. Subsequently, the model computes the right commands.

After the inference is done to compute an action, it should be transmitted to the flight controller for execution. In our work, we remap the action topic to the corresponding execution topic, to send specific message type of control commands at different levels to the PX4 controller via MAVROS.

\end{document}